\documentclass{article}
\usepackage[preprint]{neurips_2020}
\usepackage[utf8]{inputenc} % allow utf-8 input
\usepackage[T1]{fontenc}    % use 8-bit T1 fonts
\usepackage{hyperref}       % hyperlinks
\usepackage{url}            % simple URL typesetting
\usepackage{booktabs}       % professional-quality tables
\usepackage{graphicx}
\usepackage{amsfonts,amssymb,amsthm,amsmath,latexsym}     % blackboard math symbols
\usepackage{nicefrac}       % compact symbols for 1/2, etc.
\usepackage{microtype}      % microtypography

\usepackage{algorithm,algorithmic}
\theoremstyle{plain}

\usepackage{subcaption}
\newlength{\tempdima}
\newcommand{\rowname}[1]% #1 = text
{\rotatebox{90}{\makebox[\tempdima][c]{\textbf{#1}}}}

\usepackage{amsmath}
\title{Causal Structure Learning: a Bayesian approach based on random graphs}

\author{
  Mauricio Gonzalez-Soto \\
  Coordinación de Ciencias Computacionales\\
  Instituto Nacional de Astrofisica Optica y Electronica (INAOE)\\
  Mexico \\
  \texttt{mauricio@inaoep.mx}
   \And
   Ivan R. Feliciano-Avelino \\
   Coordinación de Ciencias Computacionales\\
  Instituto Nacional de Astrofisica Optica y Electronica (INAOE)\\
  Mexico \\
  \texttt{ivan.feliciano@inaoep.mx}
  \And 
   L. Enrique Sucar\\
   Coordinación de Ciencias Computacionales \\
   Instituto Nacional de Astrofisica Optica y Electronica (INAOE) \\
   Mexico\\
   \texttt{esucar@inaoep.mx}
   \And
  Hugo J. Escalante Balderas \\
  Coordinación de Ciencias Computacionales \\
  Instituto Nacional de Astrofisica Optica y Electronica (INAOE) \\ \\
   \texttt{hugojair@inaoep.mx}
}
\begin{document}

\maketitle
\begin{abstract}
A Random Graph is a random object which take its values in the space of graphs. We take advantage of the expressibility of graphs in order to model the uncertainty about the existence of causal relationships within a given set of variables. We adopt a Bayesian point of view in order to capture a causal structure via interaction and learning with a causal environment. We test our method over two different scenarios, and the experiments mainly confirm that our technique can learn a causal structure. Furthermore, the experiments and results presented for the first test scenario demonstrate the usefulness of our method to learn a causal structure as well as the optimal action. On the other hand the second experiment, shows that our proposal manages to learn the underlying causal structure of several tasks with different sizes and different causal structures.
\end{abstract}

\section{Introduction}
Intelligent agents often face situations in which an action must be chosen in the presence of uncertain conditions \citep{lake2017building}. In many real-world applications, an agent doesn't have access to all of the parameters and information required to calculate the maximum expected utility as prescribed by the von Neumann-Morgenstern and Savage Theorems \citep{von1944theory,savage1954the,gilboa2009decision}, but if the agent knew that her actions and their consequences are \textit{causally related}, then she could attempt to discover these relations and \textit{use them} in order to predict consequences of actions better than if she only observes multiple action-outcome pairs as done in Reinforcement Learning \citep{sutton1998reinforcement}.
	
Many real-world applications of decision making are solved by \textit{associative} methods which capture only statistical patterns found in data. For example, Reinforcement Learning (RL) methods, although they have good performance, such methods are not designed to explain \textit{why} a specific trajectory was chosen. This is highly relevant in real-world applications, e.g. self-driving cars, where it is very important to understand, e.g., why an accident happened. 

Causality deals with regularities found in a given environment which are stronger than probabilistic relations in the sense that a causal relation allows for evaluating a change in the \textit{consequence} given a change in the \textit{cause} \citep{spirtes2000causation}. We adopt here the \textit{manipulationist} interpretation of Causality \citep{woodward2005making}. 
The main paradigm is clearly expressed by \cite{campbell1979quasi} as \textit{manipulation of a cause will result in a manipulation of the effect}. 
Consider the following example from \cite{woodward2005making}: manually forcing a barometer to go down won't cause a storm, whereas the occurrence of a storm will cause the barometer to go down. 

Learning a causal model of an environment and using it to act upon the environment allows us to \textit{explain} aspects of the model that an associative model would not be able to: it allows one to ask \textit{why}. On the other hand, a causal model allows to \textit{imagine} what would have happened if another course of action had been taken \citep{pearl2018why}. Since a causal model, once it is learned, does not depend neither on the agent nor its preferences, the gained knowledge could be easily transferred for problems of similar domain where the use of the acquired causal relations is useful.

Given that human beings focus on \textit{local} aspects while learning causal relations which are later unified into a single structure \citep{fernbach2009causal,waldmann2008causal,danks2014unifying}, \cite{wellen2012learning} propose a model to explain how observations  and interventions are used by human beings to learn causal relations in terms of a local prediction-error learning. Following this line of thought, we propose here a local probabilistic encoding of the uncertainty that a decision maker has over the existence or not of causal relations between variables.

When doing Bayesian modeling \citep{bernardo2000bayesian,gelman2013bayesian} one first identifies the source of the uncertainty; e.g., the parameter of a probability density function which generates data; then, one specifies a probabilistic model over such uncertainty. In this work, we identify as our source of uncertainty the existence or not of a causal relationship between a given pair of variables. We will model such uncertainty as the probability of the occurrence of an edge in a random graph. 

Random Graphs were proposed by Erdös and Renyi while using probabilistic methods in order to study problems in graph theory. A random graph can be thought of as a dynamic object which starts as a set of vertices and succesive edges are added at random according to some probability law. The simplest example consists of drawing at random a graph from the space of all graphs in $n$ vertices and $M$ edges, where each graph has the same probability \citep{bollobas2001random}. Further models can be found in complex systems, economics, the study of social networks among others \citep{jackson2010social,newman2018networks}.

Our probabilistic model over the source of uncertainty, which is the decision maker's belief about the existence of causal relations between variables, is to be updated in terms of what is \textit{observed} from interactions with the environment and therefore with the true causal mechanism that controls the environment. Since observations alone do not suffice to uniquely determine a Directed Graph, we make use of interventions \citep{eberhardt2007causation}, as well as a partial ordering between the variables in the model that, at least in principle, can not be a cause of each other. This ordering must be an input product from expert knowledge.

To test our causal structure learning proposal, we tackled two tasks. In the first one, we propose that an agent learns the causal structure while taking advantage of the knowledge that she is acquiring in order to select an optimal action. In the second problem, we focus on only learning the causal structure in different configurations of a same task. The obtained results are encouraging because they show the usefulness of a Bayesian approach to learn the true underlying causal mechanism within a decision making problem. Besides, we can see that an agent can explore and exploit its causal knowledge in order to learn faster while also reaching a higher reward than if only using associative information.

\section{Related Work}
\label{related_work}
In recent work by \cite{lattimoreNIPS2016,sen2017identifying} and \cite{gonzalez2018playing}, a decision maker is given the task to learn an optimal action from a series of learning rounds in a causal environment when the causal model which controls the environment is unknown to the decision maker. \cite{lattimoreNIPS2016} assume that the conditional distributions for each variable given its parents is known, while \cite{sen2017identifying} assume as known only a part of the causal graphical model and allows for interventions in the unknown part of the model. Meanwhile, \cite{gonzalez2018playing} assume that the \textit{graphical structure} of the true model is known and attempt to learn both the parameters of the model as well as the optimal action. 

In \cite{gonzalez2018playing}, a decision maker knows the graphical structure of the true causal model which controls her environment and holds \textit{beliefs}, in form of a Dirichlet distribution, over the parameters. Following \cite{joyce1999foundations}, the decision maker forms a \textit{local} causal model and uses a result found in \cite{pearl2009causality} in order to obtain the optimal action $a^\ast$ as follows:
\begin{equation}
a^\ast=\textrm{argmax}_a \sum u(c)P(c|do(a)).
\end{equation}
Once the optimal action has been found for the current model, the decision maker observes the response from the environment, and updates the Dirichlet distribution according to what has been observed.

The methodology described above achieves a performance similar to the classical Q-Learning algorithm with the extra of learning a causal model of a very simple environment. Learning a causal model of the environment allows to extract high-level insights of a phenomena beyond associative descriptions of what is observed. A causal model is able to \textit{explain} why a particular decision was made since it allows to extract the causes and effects of an agent's actions. Once a causal model is acquired, an external user is able to reason about \textit{what...if...} statements that associative methods cannot answer.  

We note that the three mentioned works have in common that a causal graphical model controls the environment and the limitation of assuming as known a part of the causal graphical model. We propose here a random graph-based approach for causal structure learning which can be used while simultaneously learning an optimal action.

Algorithms to discover causal relations in data can be found in \cite{spirtes2000causation,eberhardt2008almost,hauser2012two,hauser2014two,hyttinen2013experiment,loh2014high,shanmugam2015learning,mooij2016distinguishing}. The most important of them being the PC-Algorithm \citep{spirtes2000causation} which is a constraint-based method for causal discovery, and has an exponential worst-case running time. The PC-Algorithm is based in the Inductive Causation Algorithm proposed by \cite{judea1990equivalence}. Other important causal discovery algorithms are the Fast Causal Inference (FCI) from \cite{spirtes2000causation} as well as the GES from \cite{chickering2002optimal}. In \cite{hauser2012two} an \textit{active learning} approach to prove a conjecture stated in \cite{eberhardt2008causal} about the worst-case number of experiments required to uniquely discover a causal graph (up to Markov equivalencies).

Also, an \textit{active learning} approach can be used to learn causal models from data, where one starts with some initial graph and then selects the instances in the data that allow to add and orient edges in order to end with a fully oriented graph. Active learning algorithms for causal discovery can be found in \cite{tong2001active,murphy2001active,meganck2006learning,he2008active,hauser2012two,10.1007/978-3-319-56970-3_9,rubenstein2017probabilistic}. Notice that these papers consider learning while interacting with a causal environment. For a recent review in causal discovery see \cite{glymour2019review}.

Our work also assumes interaction, and interventions, over an environment in which actions and outcomes are connected via a causal graphical model. On the other hand, our work does not assume an initial causal structure, rather we consider initial \textit{beliefs} about the existence of causal relations. 

\section{Methodology}
Let a rational decision maker consider the following set of variables $\mathcal{X}=\{ X_1,...,X_n \}$ which are causally related by some unknown, fixed causal graphical model (CGM) \citep{koller2009probabilistic,sucar2015probabilistic} $\mathcal{G}$; i.e., there exists some CGM $\mathcal{G}$ which has nodes corresponding to each variable in $\mathcal{X}$. The agent knows that she can only intervene variables which belong to a subset $\mathcal{X}_1 \subseteq \mathcal{X}$, and does so in order to alter the value of some  identified \textit{reward variables} $\mathcal{X}_2 \subseteq \mathcal{X}$. In the case in which $\mathcal{X}_{1,2} = \{ X_i \}$, as we will later consider in one of the test scenarios, then without loss of generality, we assume that the agent can only intervene on $X_1$ wishing to affect $X_n$. Also, we assume that the agent knows a \textit{causal ordering} of the variables, which specifies, for some but not all of the variables, which other variables can not be a cause of it. 

As mentioned in the introduction we will form a \textbf{random graph} using the beliefs that the decision maker has about the existence of a causal link between nodes in a graph. Formally, let $p_{ij} \in [0,1]$ be the \textit{belief} that the agent has over a causal relation (directed link) existing between the variable with index $i$ and the variable with index $j$. This is, the decision maker has belief $p_{ij} \in [0,1]$ that $X_i \to X_j$

Let $G$ an initial {\em random graph} formed as follows: the node set is $N=\{1,...,n\}$ and a there exists a link between $i$ and $j$ with probability $p_{ij}$. This could be thought of as throwing a charged coin with $p=p_{ij}$ and connecting nodes $i$ and $j$ if head turns up.

Now, in order to identify causal effects \citep{hauser2014two}, the agent makes an intervention $a^\ast$ over the possible values that $X_1$ can take within the resulting graph $G$. Once the action is taken, a full realization $X_1=x_1,...,X_n=x_n$ is observed. We are allowing random interventions for generality, but the random action taken could easily be replaced by an optimal action found, for example, according to the methodology found in \cite{gonzalez2018playing}. In fact we will later consider the problem of exploration vs. exploitation and propose a mix of random and optimal actions in order to learn a causal structure and the optimal action for such structure.

We recall that in \cite{gonzalez2018playing}, a decision maker knows the causal structure of the true causal model which controls her environment. The decision maker uses a Dirichlet distribution in order to model the conditional probabilities that appear in the post-interventional distribution expression obtained by Pearl's do-calculus over the known graph. A sample from the Dirichlet distribution is used to form a local causal model over the known structure in order to find the optimal action using a result from J. Pearl stated in Section 4.1 of \cite{pearl2009causality}. 

Then, in order to update the initial beliefs $p_{ij}$ according to what has been observed, we use Bayes Theorem as follows: for each pair of indexes $i,j$ and the current graph $G$, we ask for the probability of such graph producing the output $(X_1 = a^\ast, ..., X_i = x_i, ... X_j=x_j, X_n=x_n)$, which will be used as the likelihood of data given the current graph, and as a prior probability we simply use $p_{ij}$, so we have
\begin{equation}{\label{bayesian_updating}}
p_{ij}^{t+1} \propto p(X_1 = a^\ast, ... ,X_i = x_i,... ,X_j=x_j, ... X_n=x_n | \textrm{current graph})p_{ij}^t.
\end{equation}
Then, we update the model generating a new graph according to $p_{ij}$.

%\subsection{Implementation}
%\begin{enumerate}
%\item Initialize $p_{ij}$ randomly
%\item Initialize graph $G$ with link from node $i$ to node $j$ with probability $p_{ij}$
%\item Take action $a^\ast$ for graph $G$ and probabilities given by a count of the observations.
%\item Update $p_{ij}^{t+1} \propto p(X_1 = a^\ast, ... , X_i = x_i, ..., X_j=x_j, ... , X_n=x_n | \textrm{current graph})p_{ij}^t.$
%\end{enumerate}

\begin{algorithm}
\caption{Causal relations belief learning algorithm}
\begin{algorithmic}
    \REQUIRE{The maximum number of iterations $k$.}
    \STATE{Initialize $p_{ij}$ randomly.}
    \FOR{$t = 1, \dots, k$} 
    \STATE{Initialize $G$ with link node $i$ to node $j$ with probability $p_{ij}$.}
    \STATE {Take action $a*$ for graph $G$ and probabilities given by a count of the observations.}
    \STATE{Update $p_{ij}^{t+1} \propto p(X_1 = a^\ast, ... , X_i = x_i, ..., X_j=x_j, ... , X_n=x_n | \textrm{current graph})p_{ij}^t.$}
    \ENDFOR
    \RETURN{The beliefs $p_{ij}$}
\end{algorithmic}
\end{algorithm}

\subsection{Exploration vs. Exploitation}
If an optimal action is selected, then the decision maker takes the risk of not observing enough events in order to update his beliefs due to the exploration vs. exploitation tradeoff \citep{march1991exploration}. We propose to use an $\varepsilon$-greedy method \citep{sutton1998reinforcement} with a linearly decaying and exponentially decaying $\varepsilon$.

\section{Experiments}
Our objective is to show that the structure of the causal model which controls an environment can be learned via repeated interaction by a rational decision maker. We consider two problems: the disease-treatment scenario proposed by \cite{gonzalez2018playing} and the multistep task of light switch control introduced by \cite{nair2019causal}.

We carried out a series of experiments in which an agent, intervening one variable at a time, updates its causal relations beliefs until they converge to a value that corresponds to whether the connection exists or not. For true connections, the beliefs must reach a high value, for spurious ones a very low value must be reached. These values correspond to the degree of certainty of the decision maker over the existence of a causal relation between pairs of variables.

\subsection{Disease treatment}

We examined the hypothetical example proposed by \cite{gonzalez2018playing}: consider a patient who can have one of two possible diseases. A doctor can treat the disease with either treatment $A$ or $B$, both of which carry some risk.
Whether a patient is cured or not depends on the disease the patient has, the given treatment and a possible adverse reaction that the latter may have on the patient. Having said that, we can define the set of binary variables for the problem, the interve variables, and the reward variables as : $\mathcal{X} = \{Treatment, Disease, Reaction, Lives\}$, $\mathcal{X}_1 = \{ Treatment\}$  and $\mathcal{X}_2 = \{Lives\}$, respectively.

\begin{figure}
    \centering
    \includegraphics[scale=0.35]{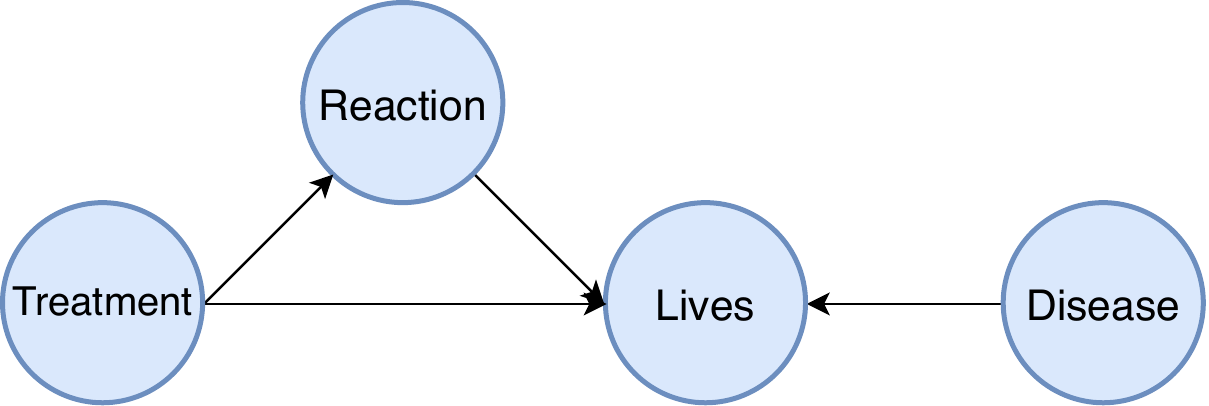}
    \caption{Causal structure underlying the disease-treatement problem.}
    \label{fig:disease-treatment-struct}
\end{figure}

We propose to mimic the physician-patient interaction with an agent interacting with an environment which is ruled by a causal model. The structure of the causal model is shown in Figure \ref{fig:disease-treatment-struct}.

\subsubsection{Implementation}

% In general, our method consists of two stages: 1) an exploration for data generation, and 2) an online updating of the beliefs.

% The exploration stage comprises randomly select a treatment during $L$ steps and saving the values of the remain variables as an observation of the environment state in a buffer $O$. 
% The decision maker  generates a graph $G$ based on its beliefs, where each 
% belief $p_{ij}$ is initialized with a value of $0.5$.

% Therefore, at every learning time step, for each of $p_{ij}$, where $(i, j)$ denotes a directed edge from variable $i$ to $j$, we compute a random number $r \in [0,1]$ such that if  $r \leq p_{ij}$ then the directed edge $(i, j)$ is added to the set of arcs in $G$.	

% Because of the problem essence, it is possible to limit the agent search space of the possible causal mechanisms and avoid generating cycles. We use the causal order defined as $(Disease, Treatment, Reaction, Lives)$ to limit the beliefs and hence the feasible edges in $G$.

% Then, with the data yielded from an exploration stage 
% the agent estimates the conditional probability 
% distribution for each variable using the Maximum 
% likelihood estimation method, such that they be 
% consistent with the induced graph $G$ from the beliefs. With this information, the agent has an approximate model consisting of a causal structure and its parameters.

% The beliefs updating is done at every interaction with 
% the environment.

The set of beliefs, to generate the graph $G$, is restricted to the possible edges that follow the causal order
$(Disease, Treatment, Reaction, Lives)$.
For this problem, the only variable that can be intervened is 
\textit{Treatment}, i.e., $\mathcal{X}_1 =\{ Treatment\}$.
We propose to use and compare three approaches to set the treatment. The first action selection policy is 
not focused on exploiting causal knowledge to select the best treatment, so the agent randomly selects it.
The remaining two, try to deal with the exploitation and exploration
problem through an $\epsilon$-greedy policy with an exponential and a linear $\epsilon$ decay, respectively.
% At each training time step $t$, the agent sets the value of 
% treatment and wait for Nature response. 
Once a treatment is performed, the agent observes the environment 
response and adds it to an observations buffer $O$. Therefore, the agent carries out an update of its beliefs using \ref{bayesian_updating}.
The parameters of the models are learned using a Maximum likelihood estimator, a Laplacian smoothing, and the data $O$, collected from the beginning of learning up to the current learning time step.
% \[
% 	p^t_{ij} \leftarrow \frac{p^{t-1}_{ij} \times P_{ij}(o)}{p^{t-1}_{ij} \times P_{ij}(o) + (1 - p^{t-1}_{ij}) \times P_{\lnot ij}(o)}.
% \]

% To compute each joint probability $P_{ij}$ and $P_{\lnot ij}$, the agent must 
% estimate the parameters of two models.
% The first model corresponds to the current graph $G$ with the edge
% $(i, j)$ and the second model corresponding to the $G$ without 
% the edge $(i, j)$. The edge $(i, j)$ is added or removed according to the case.

\subsubsection{Parameters}

% As mentioned earlier the set of variables for our problem $\{Treatment, Disease, Reaction, lives\}$ and the possible edges of beliefs is restricted to folllow the causal order  $(Disease, Treatment, Reaction, Lives)$.
To validate our approach we ran $10$ times our method
to learn the causal relationships shown in Figure \ref{fig:disease-treatment-struct}. 
The initial values for beliefs $p_{ij}$ are set to $0.5$.
For each execution, the number of beliefs updating rounds is $50$. For the random
action selection policy, the probability of choosing one of the treatments is $0.5$.
For the $\epsilon$-greedy policy, the optimal reward for every 
interaction is set to $1$ which can be interpreted as that the patient lived. 
Thus, the agent is expected to select the treatment most likely to cause the patient to live.
Both versions of $\epsilon$-greedy, start with an exploration probability equal to $1$ until reaching 
$0.01$. For the exponential decay, de value of $\epsilon$ is decreased by multipling it by
$0.9$ at every step. On the other hand, the linear decay is done by uniformly decreasing $\epsilon$  with respect to the relationship between the number of interventions and the difference between the maximum and minimum exploration values.

\subsubsection{Results}

To measure the performance of our algorithm we wish to know 
how different the beliefs and ground truth (defined in Figure \ref{fig:disease-treatment-struct}) are. We use the $l^2$
norm where we directly compare the values of the beliefs 
with the true edges. As stricter measures we use the Hamming loss 
and the accuracy. Since these measures require comparing whether two
values are equal or not, before the evaluation, we propose mapping 
the beliefs to values in $\{0, 1\}$, assigning the value of $0$ to 
$p_{ij}$ if $p_{ij} < \theta$ or $1$ othewise ($\theta = 0.75$).
We evaluate each of the mentioned metrics
per episode and the average is obtained by execution.
Besides the metric to evaluate how well does our agent learn the true causal
structure, we want to know how good is at choosing a treatment. Therefore,
we compare our method following the $\epsilon$-greedy policy variants with the 
Q-learning algorithm following the same $\epsilon$-greedy versions.

\begin{figure}[h]
\centering
        \begin{subfigure}[b]{0.3\textwidth}
                \centering
                \includegraphics[width=.7\linewidth]{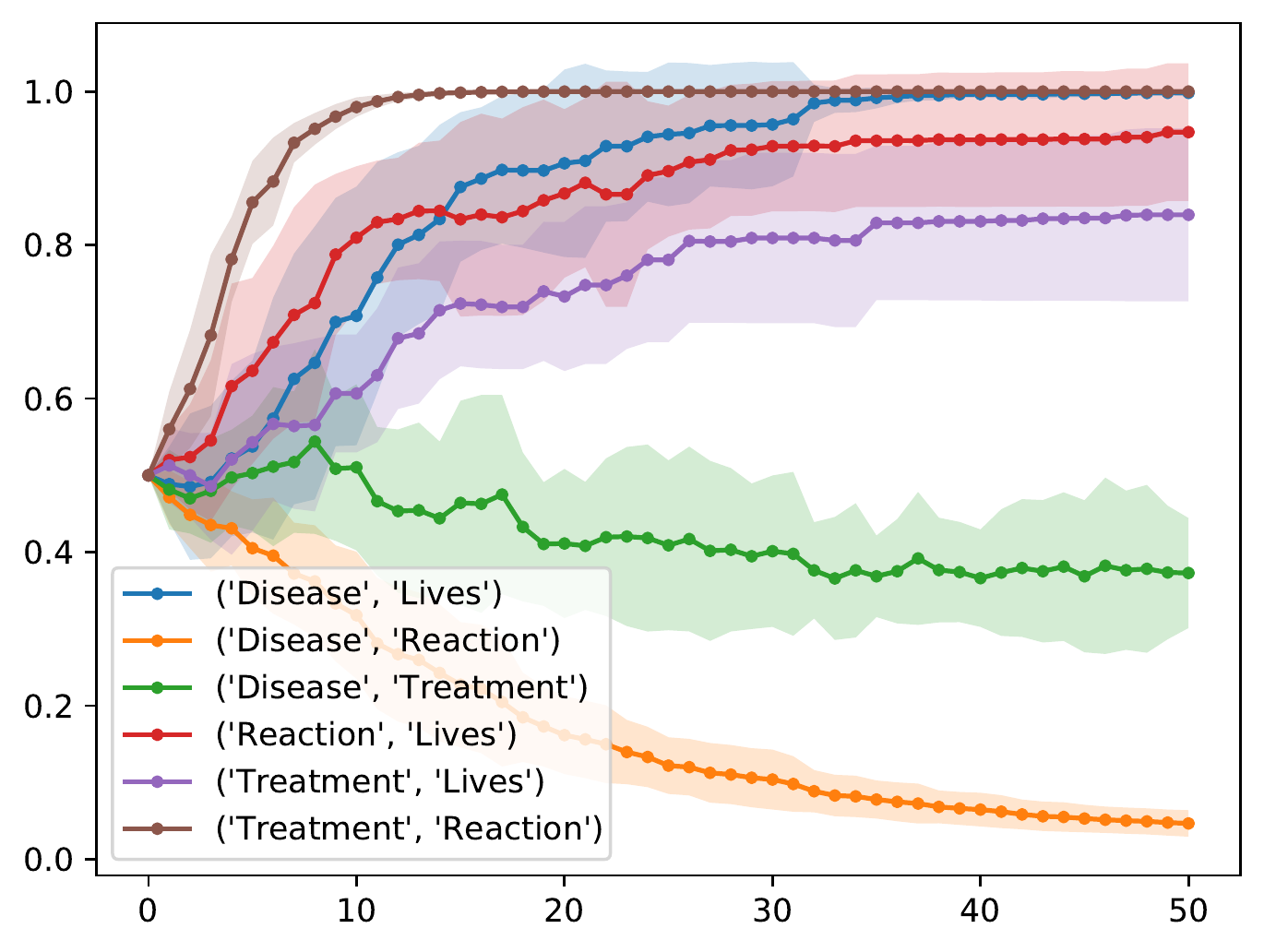}
                \caption{\scriptsize{Random policy.}}
                \label{fig:random-beliefs-disease-treatment}
        \end{subfigure}%
        \begin{subfigure}[b]{0.3\textwidth}
                \centering
                \includegraphics[width=.7\linewidth]{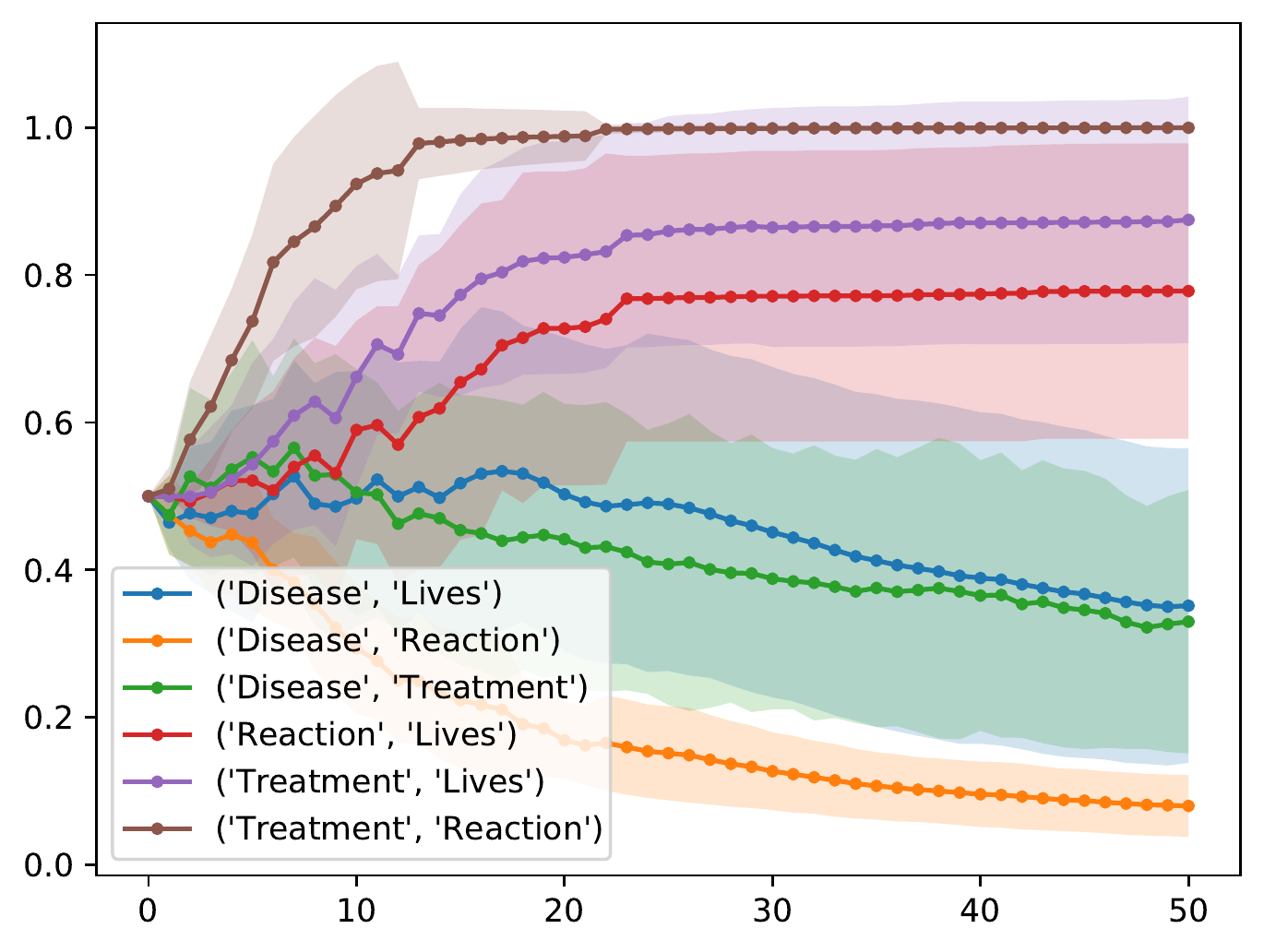}
                \caption{\scriptsize{$\epsilon$-greedy w/ exponential decay.}}
                \label{fig:egreedy-exp-beliefs-disease-treatment}
        \end{subfigure}
        \begin{subfigure}[b]{0.3\textwidth}
                \centering
                \includegraphics[width=.7\linewidth]{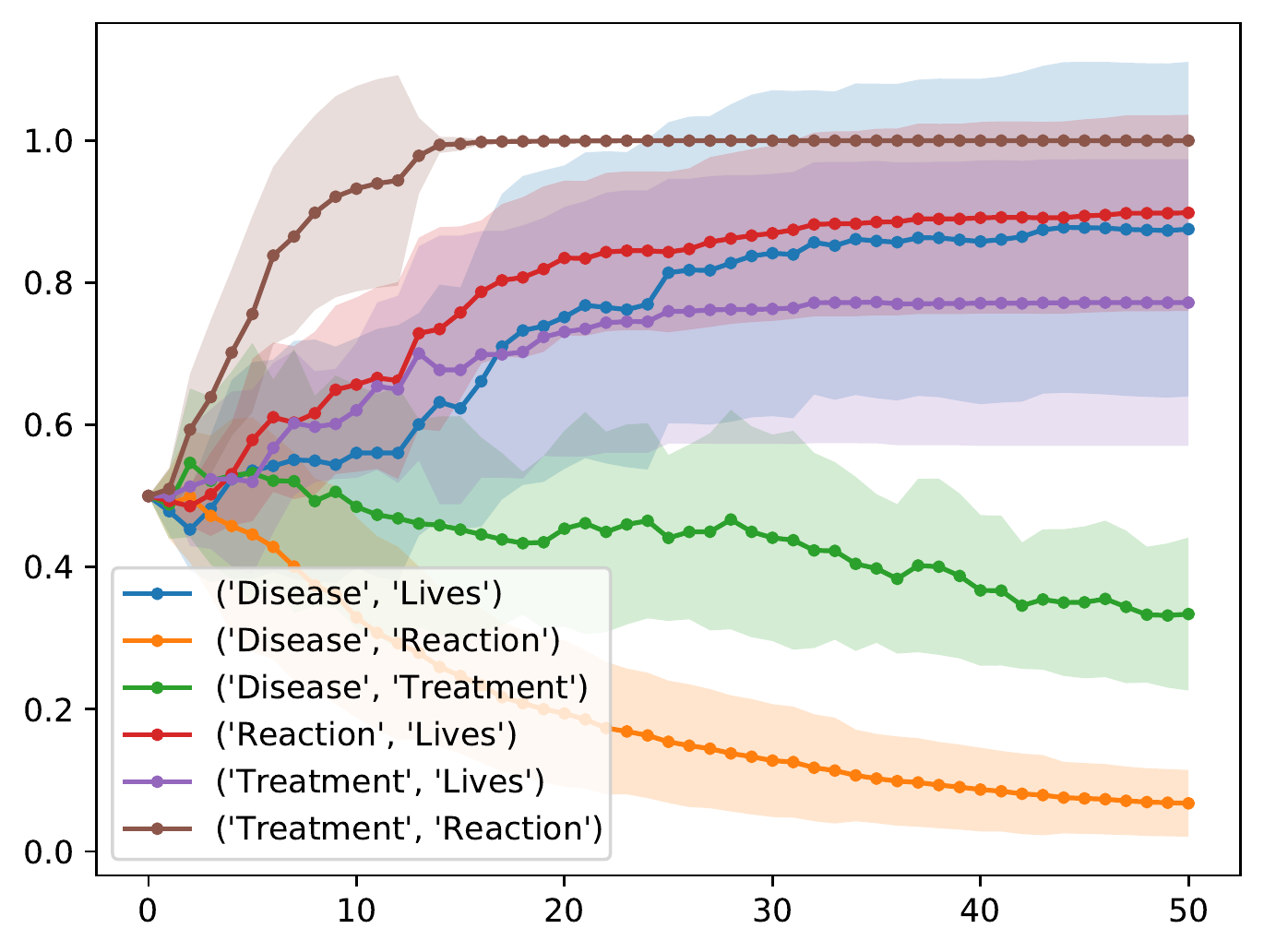}
                \caption{\scriptsize{$\epsilon$-greedy w/ linear decay.}}
                \label{fig:egreedy-linear-beliefs-disease-treatment}
        \end{subfigure}
        \caption{Average beliefs $p_{ij}$ over $50$ rounds and $10$ experiments.}\label{fig:beliefs-disease-treatment}
\end{figure}

Figure \ref{fig:beliefs-disease-treatment} shows the average value and standard deviation per round of each belief $p_{ij}$ over $10$ runs for each action policy. It is easy to observe that the relation between \textit{Treatment} and \textit{Reaction} is the easiest to learn for the three policies. 
 Also, we can see that the beliefs about \textit{Treatment}- \textit{Lives}, and \textit{Disease}-\textit{Reaction} remain very similar in all policies. However, 
there is a different behavior for \textit{Reaction}-\textit{Lives} and \textit{Disease}-\textit{Lives}. The former belief,
keeps its value much lower for the $\epsilon$-greedy with an exponential decay compared two the
other action selection policies. Something similar happens to the \textit{Disease}-\textit{Lives}, but in this case, it seem that the agent is learning the wrong way. From this, we can assume that rather than stop performing random actions and select the best, sometimes seems better to stop updating the beliefs or maybe use another method to calculate the parameters of the models to compute the new values of $p_{ij}$.

\begin{figure}[h]
        \begin{subfigure}[b]{0.225\textwidth}
                \centering
                \includegraphics[width=.85\linewidth]{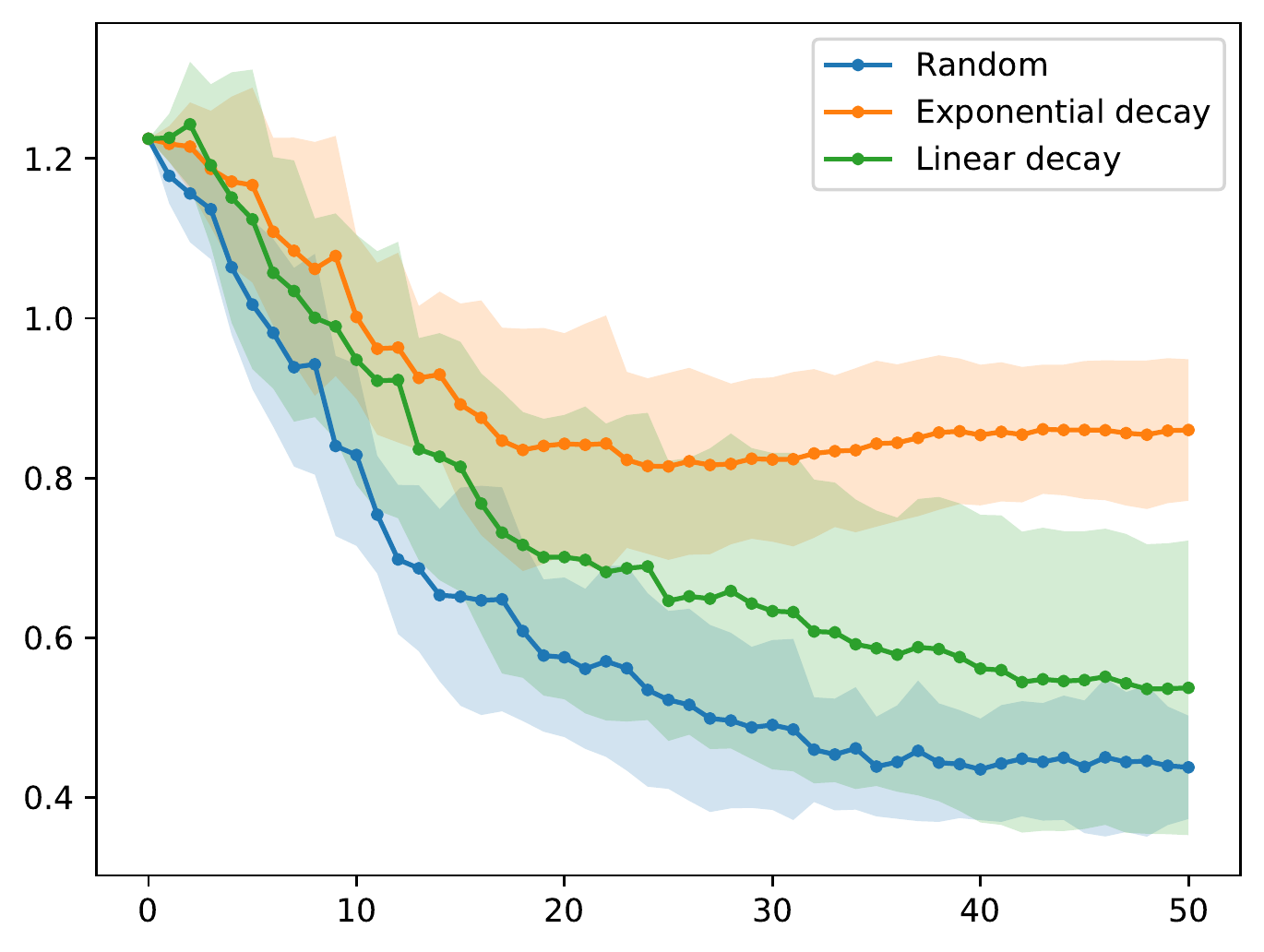}
                \caption{$l^2$ norm}
                \label{fig:l2-disease-treatment}
        \end{subfigure}%
        \begin{subfigure}[b]{0.225\textwidth}
                \centering
                \includegraphics[width=.85\linewidth]{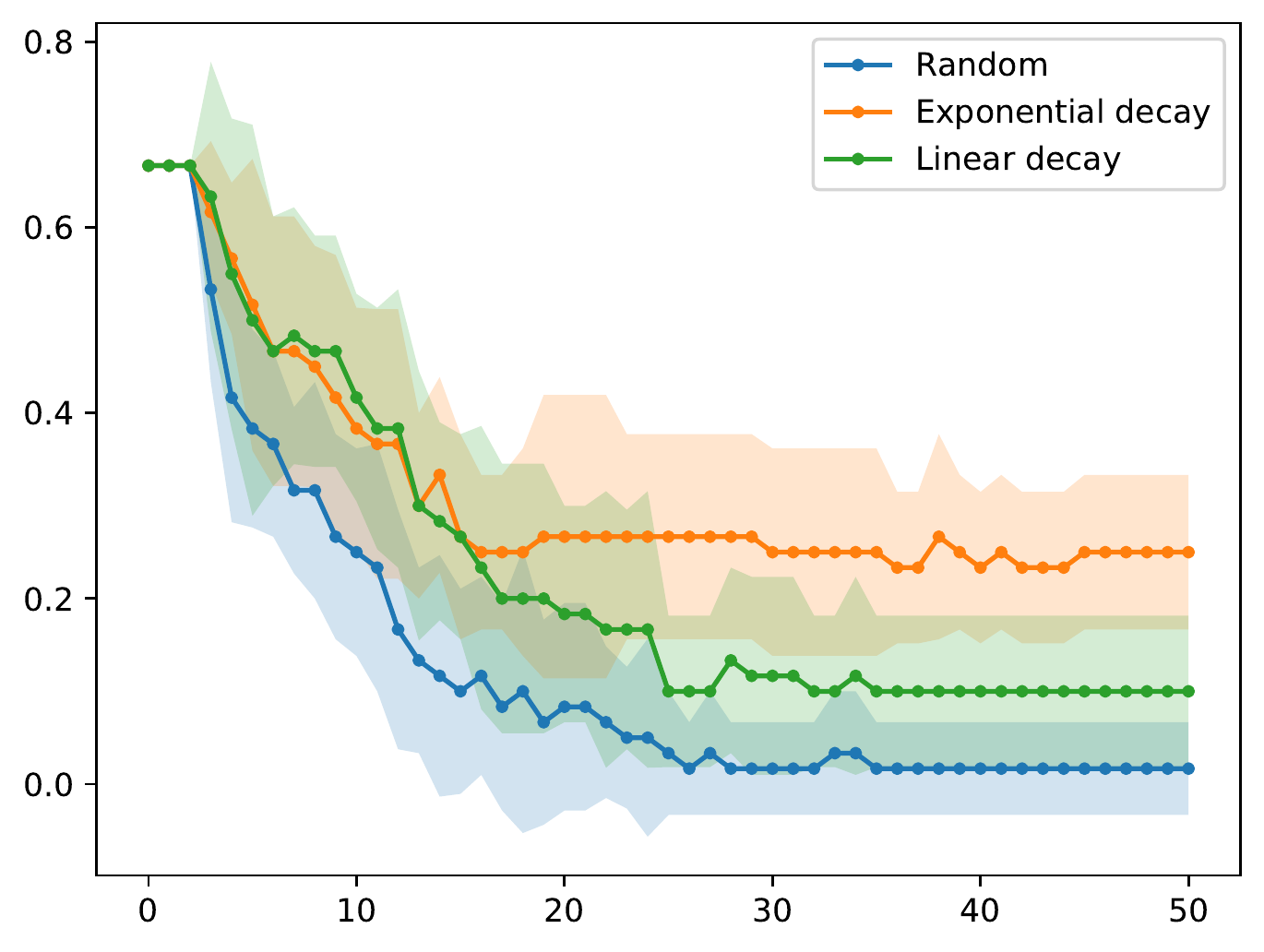}
                \caption{Hamming loss}
                \label{fig:hamming-disease-treatment}
        \end{subfigure}%
        \begin{subfigure}[b]{0.225\textwidth}
                \centering
                \includegraphics[width=.85\linewidth]{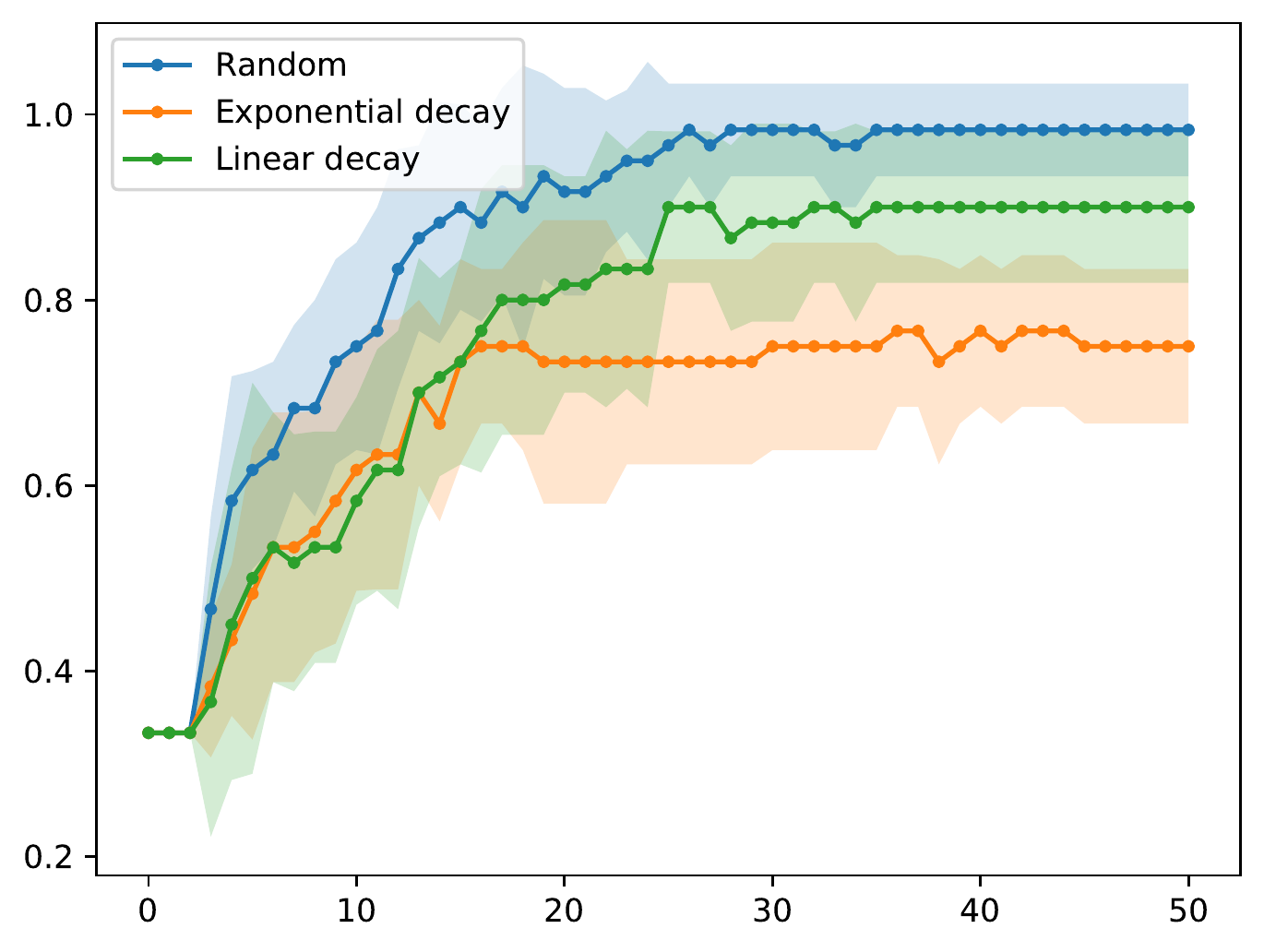}
                \caption{Accuracy}
                \label{fig:acc-disease-treatment}
        \end{subfigure}
        \begin{subfigure}[b]{0.225\textwidth}
                \centering
                \includegraphics[width=.85\linewidth]{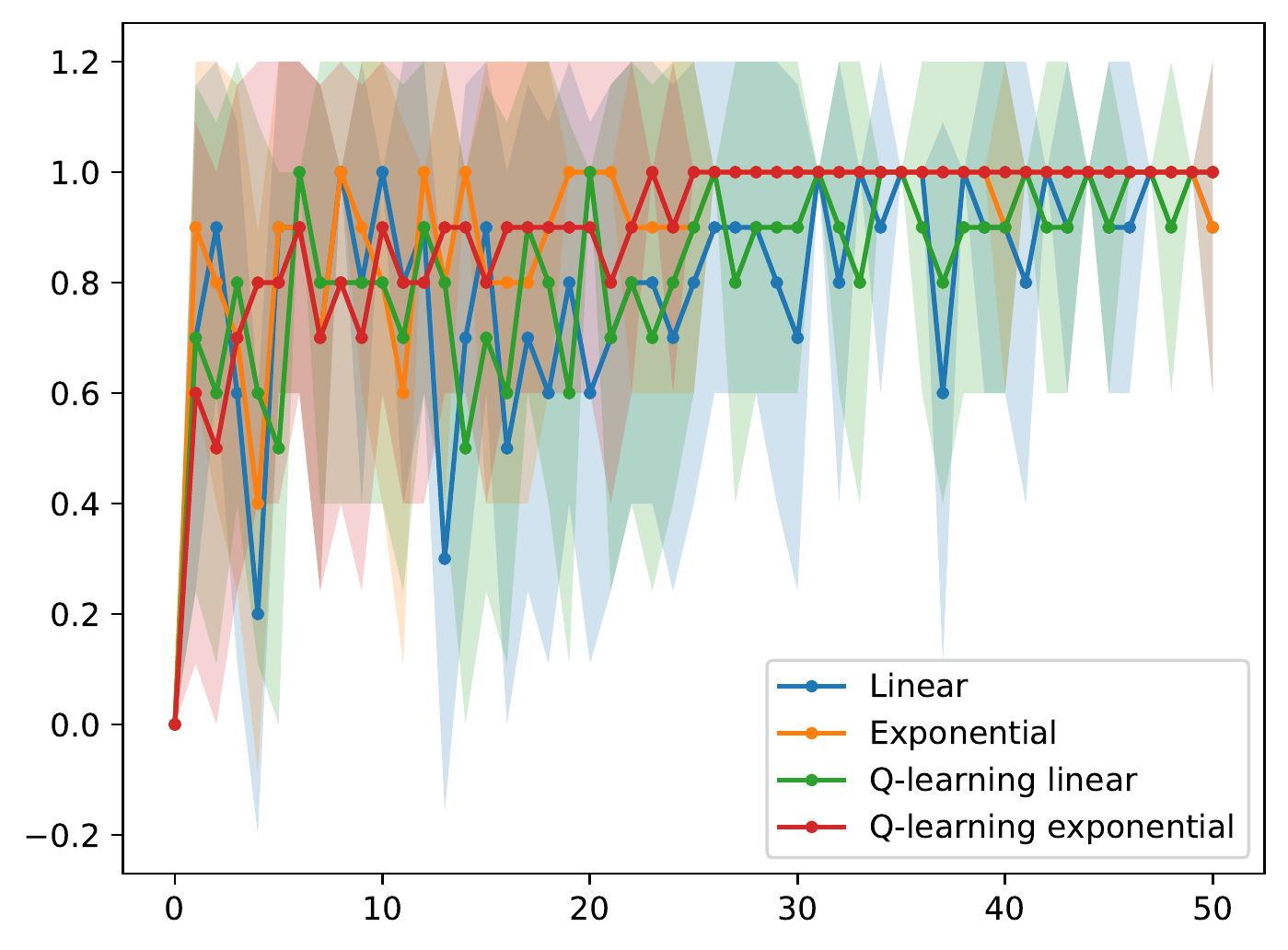}
                \caption{Reward}
                \label{fig:reward-disease-treatment}
        \end{subfigure}
        \caption{Evaluation metrics per interaction round over $50$
        rounds and $10$ experiments.}\label{fig:metrics-disease-treatment}
\end{figure}

Figure \ref{fig:metrics-disease-treatment} shows the three evaluation metrics for the structure learning and the obtained reward for each interaction round. 
The first three plots of Figure \ref{fig:metrics-disease-treatment} evidence that random actions
are better to find the true causal relationships. On the other hand, we can see that policy using a fast decay
of the exploration rate, outperforms the rest of the methods and is very similar to the Q-learning algorithm with
the same action selection scheme. However, our approach learns to choose from causal mechanisms of the world.

% \begin{table}[h]
% \centering
% \caption{Average and standard deviation for each belief at the final round over 10 experiments.}
% \label{tab:final-disease-beliefs}
% \begin{tabular}{@{}lll@{}}
% \toprule
% Directed edge & Ground truth & Belief \\ \midrule
% $Disease \rightarrow Lives$ & $1$ & $1.00 \pm 0.00$ \\
% $Disease \rightarrow Reaction$ & $0$ & $0.07 \pm 0.02$ \\
% $Disease \rightarrow Treatment$ & $0$ & $0.40 \pm 0.19$ \\
% $Reaction \rightarrow Lives$ & $1$ & $0.84 \pm 0.12$ \\
% $Treatment \rightarrow Lives$ & $1$ & $0.91 \pm 0.15$ \\
% $Treatment \rightarrow Reaction$ & $1$ & $1.00 \pm 0.00$ \\ \bottomrule
% \end{tabular}
% \end{table}

\subsection{Light switches scenario}

\begin{figure}[h]
    \centering
    \includegraphics[scale=0.17]{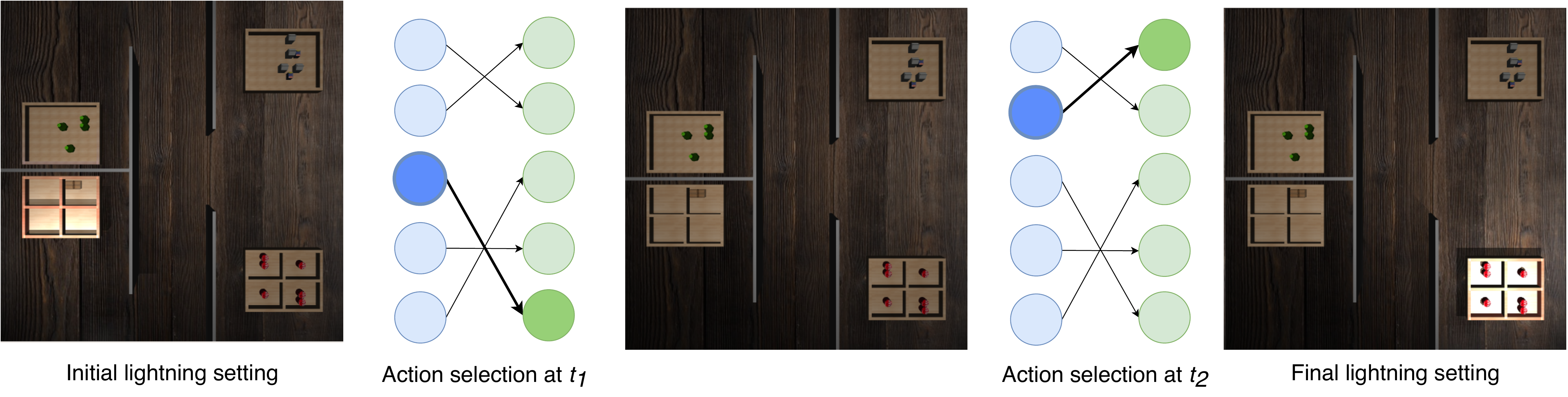}
    \caption{The agent starts with an initial configuration of the lights. It aims to reach the goal state of lights by controlling the switches at every step $t$. The connections between switches and lights can be represented by a causal model (the graph with the blue and green vertices).}
    \label{fig:light-switch-overview}
\end{figure}

We tested our approach on the light switch control tasks testbed introduced by \cite{nair2019causal}. In general, an agent starts with an initial lighting setting of the environment and aims to reach a specific configuration of lights on and off. An example of an initial observation an the goal of the agent is shown in Figure \ref{fig:light-switch-overview}.

Specifically, an agent has control of $N$ light switches to control $N$ lights. The relationship
between switches and lights is given by a causal
model which defines how the former control the latter.
The underlying causal model is unknown to the agent. However, the agent can induce the causal
structure and the model parameters through interactions with its environment, i.e., by moving the light switches.
Following the game-based approach, the decision maker performs an action (flipping a switch) and its world responds with the state of the lights and switches. 

The agent can perceive two types of signals, an image with a bird's eye view of the environment (as in Figure 
\ref{fig:light-switch-overview}) and a value for a set of binary variables $\mathcal{X} = \{X_1, \dots, X_n\}$ where $X_i\in\{0, 1\}$ encodes the state of the lights and switches. The number $N$ of switches and lights is equal, and we can define another variable which represents an action of not flipping any switch. Therefore, the cardinality of $|\mathcal{X}| = n$ is equal to $2N + 1$.
The set of variables to intervene $\mathcal{X}_1 = \{X_1, \dots, X_{N+1}\}$ describe the switches condition and the non action 
decision, where each $X_i$
denotes with $X_i=1$ if the switch has been flipping into the on position or $X_i=0$ otherwise.
Furthermore, the set of reward variables $\mathcal{X}_{2} = \{X_{N+2}, \dots, X_{2N + 1}\}$ defines the ligths state, where $X_j = 1$ indicates that the light $j$ is on, otherwise $X_j = 0$. 
% The last element of the agent observation is a variable $Y$ that indicates whether the light was turned on without explicitly indicating what it was.

\begin{figure}
    \centering
    \includegraphics[scale=0.17]{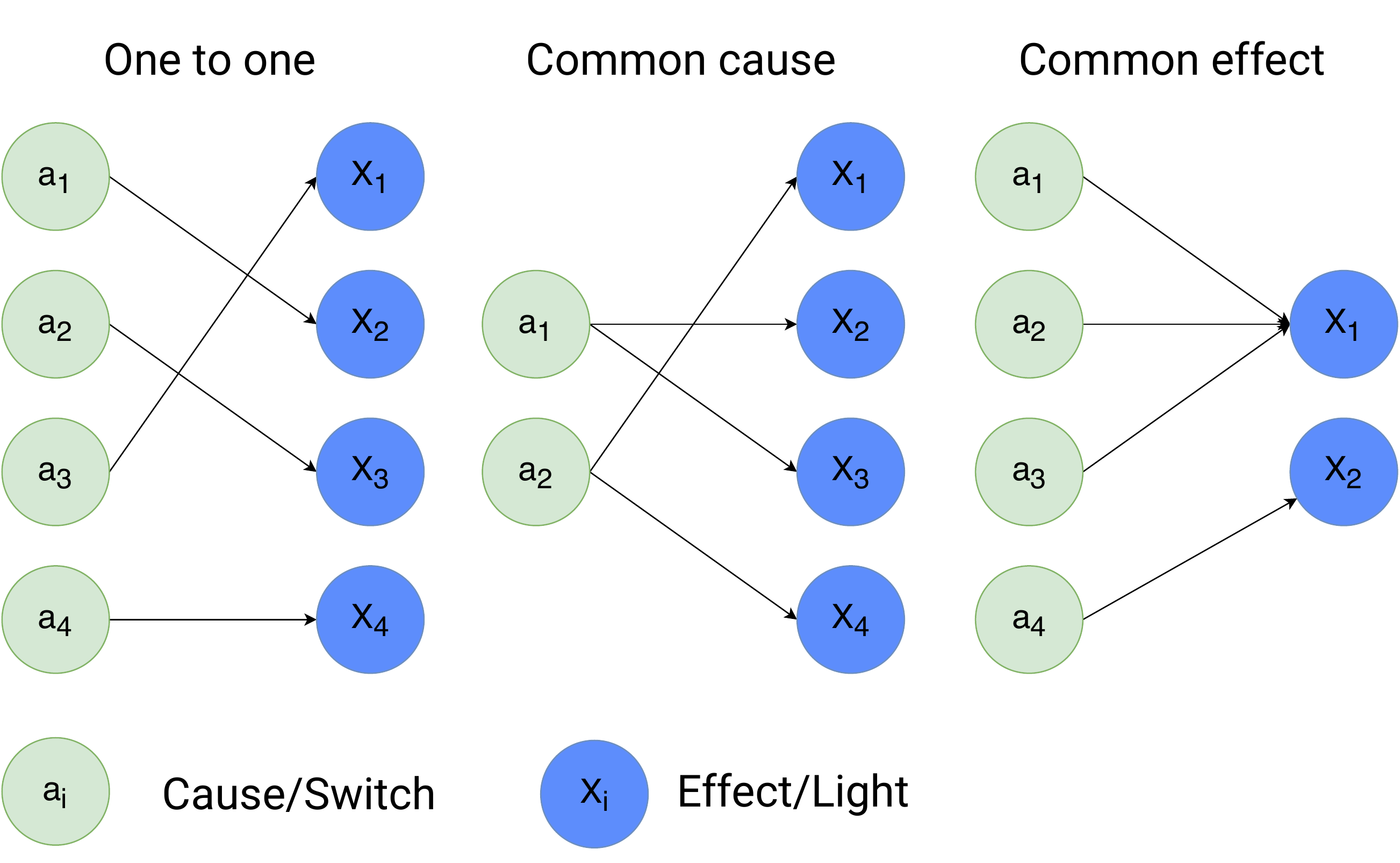}
    \caption{Examples of the 3 types of latent causal structures on the environment.}
    \label{fig:light-switches-topos}
\end{figure}

The testbed involves 3 different types of causal relationships between the switches and lights and they are shown in Figure \ref{fig:light-switches-topos}:

\begin{itemize}
    \item \textit{One to one}: each switch controls only one light. 
    \item \textit{Common cause}: in this case,
    a single switch may control more than one light and each light is controlled by at most one switch.
    \item \textit{Common effect}: each switch maps to one light, but more than one switch can control the same light.
    % \item \textit{Causal chain}: there is one switch that needs to be activated so the causal effects of the others can be observed.
\end{itemize}

\subsubsection{Implementation}

The same steps are followed as for the problem attacked above, with some exceptions and changes. First, the beliefs are restricted to $\mathcal{X}_1 \times \mathcal{X}_2$, where the sets of variables corresponding to causes and effects are $\mathcal{X}_1$ and $\mathcal{X}_2$, respectively.
Each belief $p_{ij}$ is initialized with a value of $0.5$, where
$i$ denotes the cause/switch $X_i$ and $j$ denotes the effect/light
$X_j$.
% Furthermore, we can assume that for all the cases where the structure is not
% a causal chain, the causal graph $G$ generated from the beliefs is bipartite.

Unlike the previous problem, we need to model
the conditional probability from a different way.
We model de conditional probability with two distributions. One to model the probability of turn on the light
and another one corresponding to the turn off of the
lightning. The problem is deterministic,
and for this proof of concept we do not change
this specification. So we model the probabilities with values of 1's or 0's depending of the problem instance.

We keep two buffers of observations, $O_{on}$ and $O_{off}$.
The first one, $O_{on}$ holds the data generated on interactions where the agent turns a light on.
Conversely, $O_{off}$, stores the data after the agent acts and turns off some light.
Using both, the generated data and $G$, the agent 
estimates the parameters of two models using the Maximum 
likelihood estimation method and a Laplacian smoothing. Therefore, after the agent's action selection, if some light is turned on,  we can compute the probability of
an observation using the parameters of the % lightning of 
lights on model.
Otherwise we use the parameters of the lights off model.

During the interaction for updating the beliefs, 
the agent acts by randomly fixing the value of one of the variables in $\mathcal{X}_1$, i.e., the decision maker flip the switch to the on or off position.
Therefore, at the time step $t$, after action and environment's response, the agent updates its beliefs following \ref{bayesian_updating}. 
% rules:

% \[
% 	p^t_{ij} \leftarrow 
% 	\begin{cases}
% 	\frac{p^{t-1}_{ij} \times P_{ON_{ij}}(o)}{p^{t-1}_{ij} \times P_{ON_{ij}}(o) + (1 - p^{t-1}_{ij}) \times P_{ON_{\lnot ij}}(o)} & \mbox{if a light turned on}\\
% 	\frac{p^{t-1}_{ij} \times P_{OFF_{ij}}(o)}{p^{t-1}_{ij} \times P_{OFF_{ij}}(o) + (1 - p^{t-1}_{ij}) \times P_{OFF_{\lnot ij}}(o)} & \mbox{if a light turned off}.\\
% 	\end{cases}
% \]

% The distributions $P_{on}$ and $P_{off}$, are computed with the data corresponding $O_{on}$ and $O_{off}$ and the 
% graph $G$ adding or removing the edge $(i, j)$
% according to the case.

For this proof of concept we did not use visual observations of the environment. However, our method can be extended if we encode the images as the variables of $\mathcal{X}$.

\subsubsection{Parameters}

We carried out a series of experiments to show that our proposal is able to
learn the causal structure of several instances of the task.
An experiment consists of running our method over $K = 10$ instances of the light control task where each problem instance has an undelying causal structure that guides the connections between switches and lights. We only learn structures with $N$ lights and $N+1$ posible actions, because we include the action of doing nothing. Then, the number of possible edges is $(N+1) \times N$ ($N \in \{5, 7, 9\}$).
% The number of exploration steps $L$ before the updating of
% the beliefs is $20$. 
The number of updating rounds $k$ is $50$. We cope with the three types of causal
structures: 1) one to one, 2) common cause and 3) common effect.

\subsubsection{Results}

Unlike the previous problem, we have several possible configurations for an instance of the task. Furthermore, each algorithm execution
corresponds to learn a different causal structure.
Like before, we use the $l_2$
loss function, the Hamming loss and the accuracy. For the
latter measures we set $\theta = 0.75$.

\begin{figure}
\settoheight{\tempdima}{\includegraphics[width=.2\linewidth]{example-image-a}}%

\centering\begin{tabular}{@{}c@{ }c@{ }c@{ }c@{}}
&\scriptsize{$l^2$ norm} & \quad \scriptsize{Hamming loss} & \quad \scriptsize{Accuracy} \\
\rowname{\scriptsize{$N = 5$}}&\settoheight{\tempdima}{\includegraphics[width=.2\linewidth]{example-image-a}}%

\includegraphics[width=.2\linewidth]{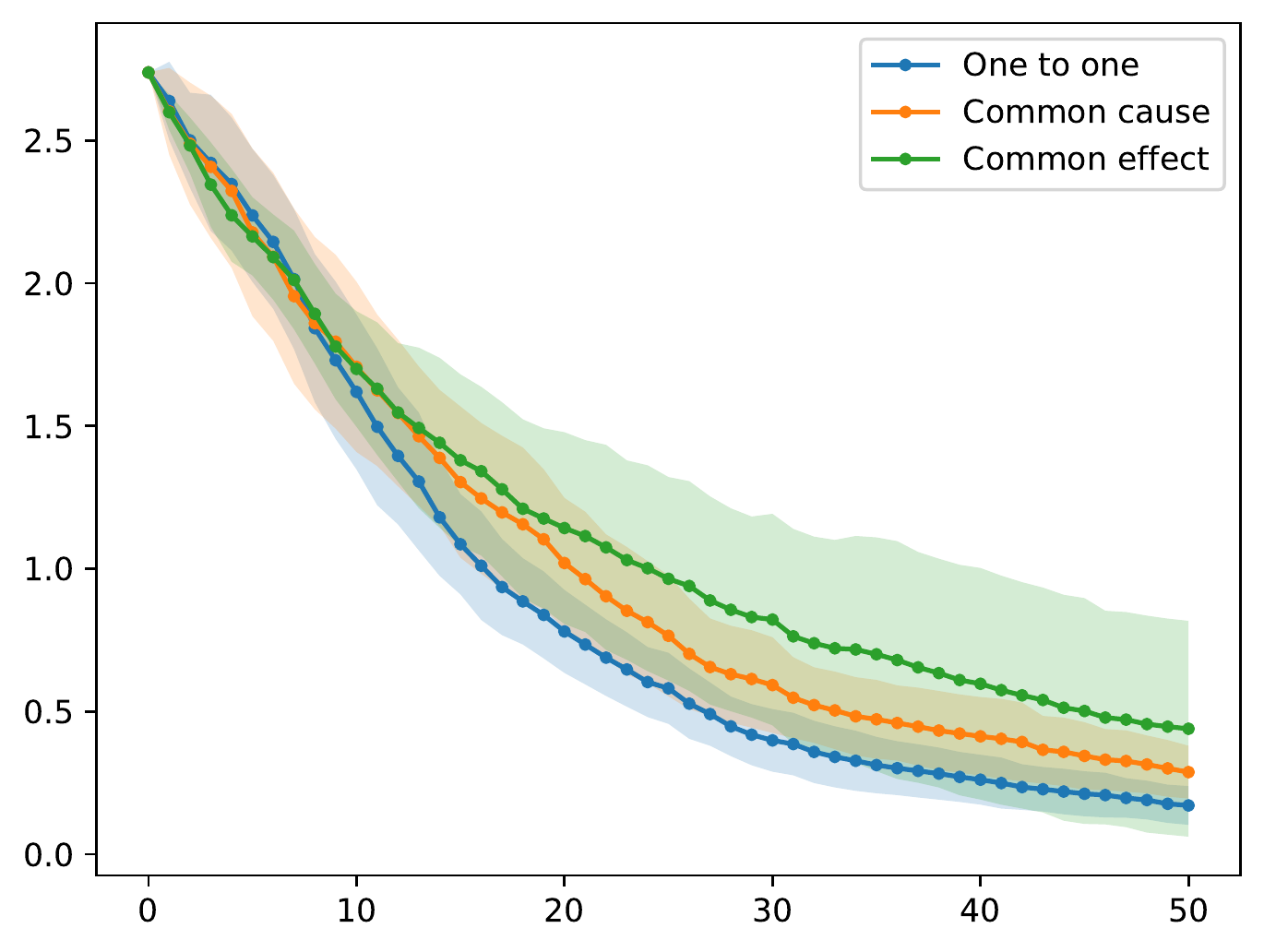} & \quad
\includegraphics[width=.2\linewidth]{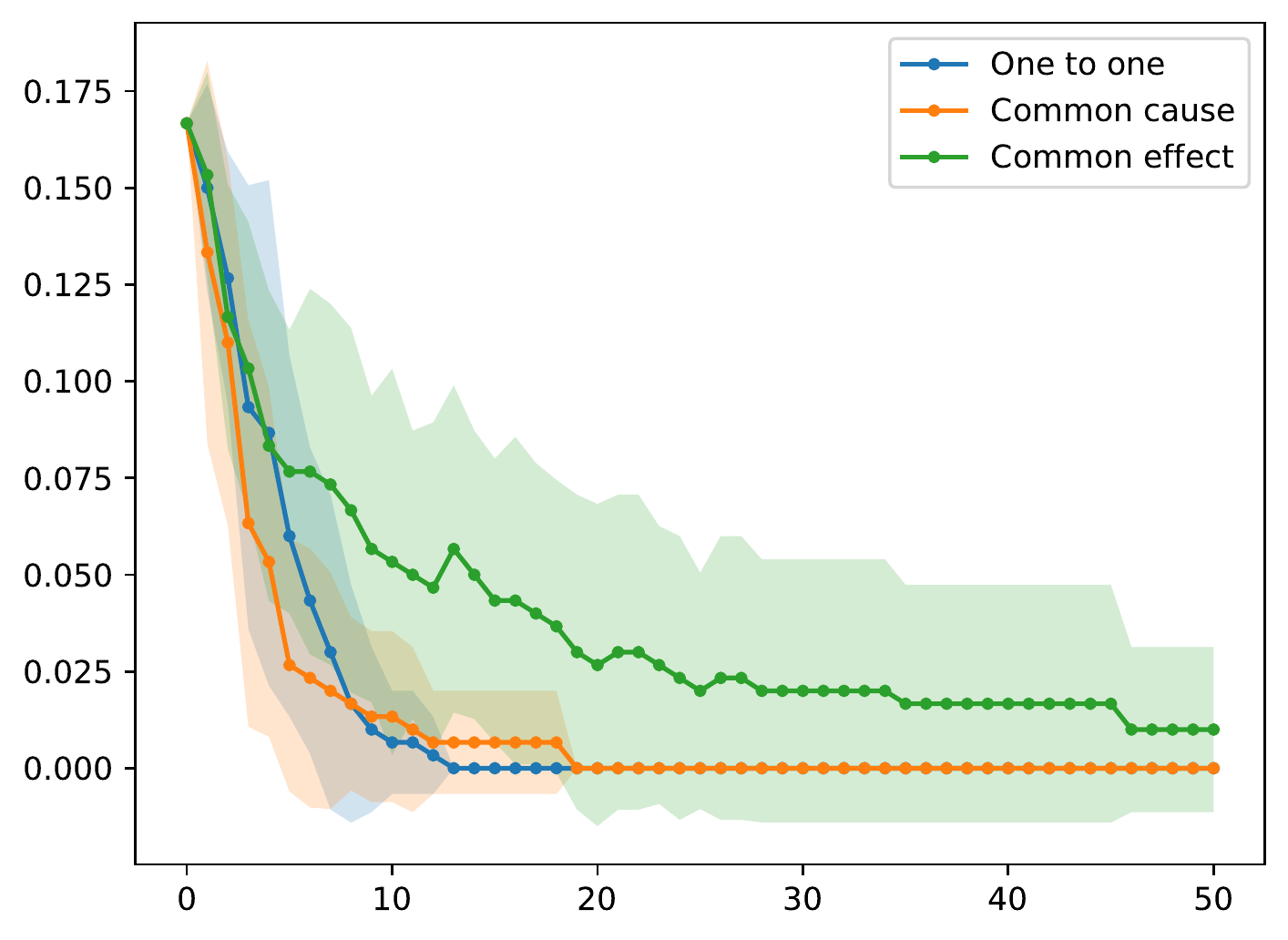}& \quad
\includegraphics[width=.2\linewidth]{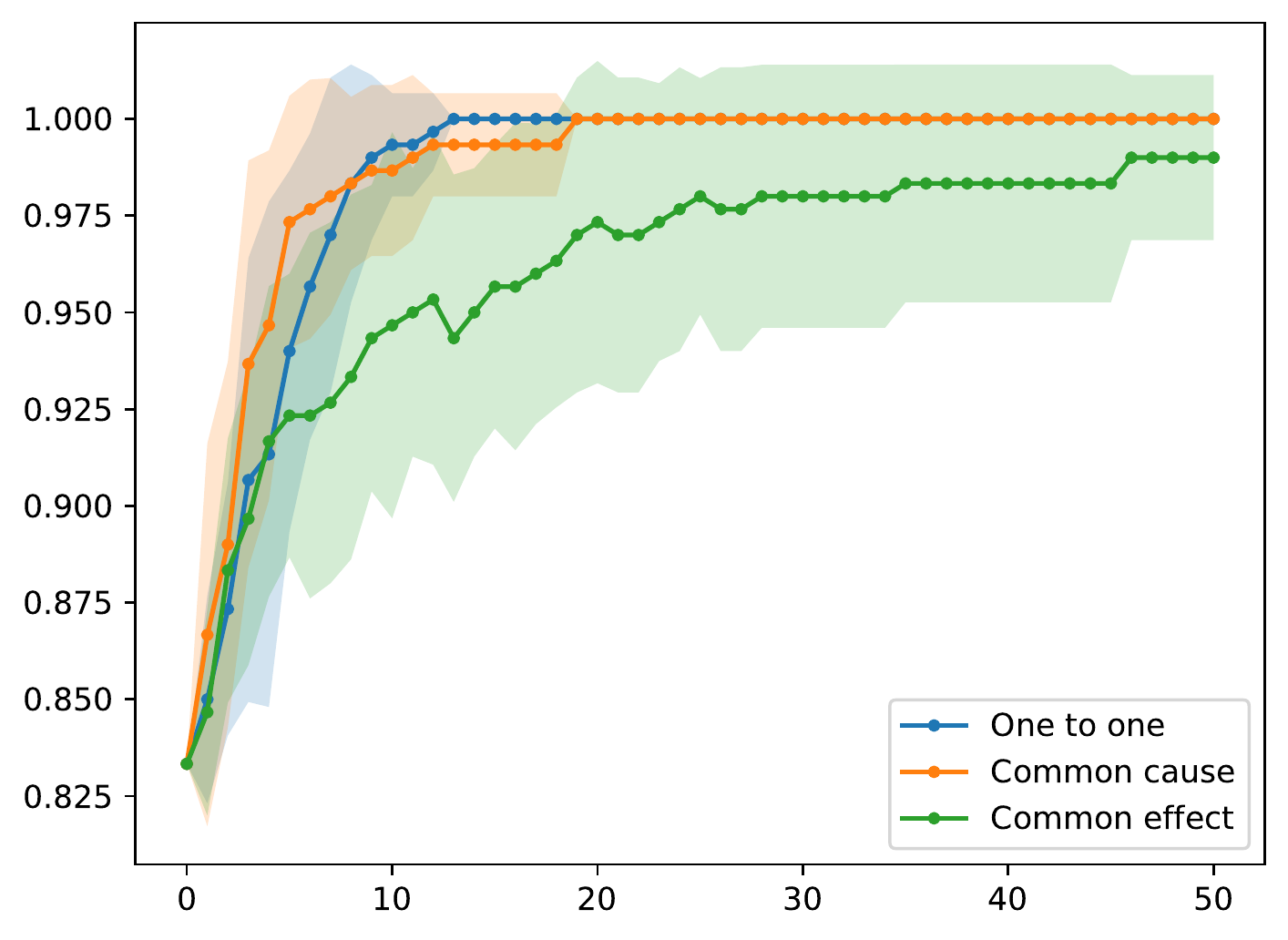}
\\
\rowname{\scriptsize{$N=7$}}&
\includegraphics[width=.2\linewidth]{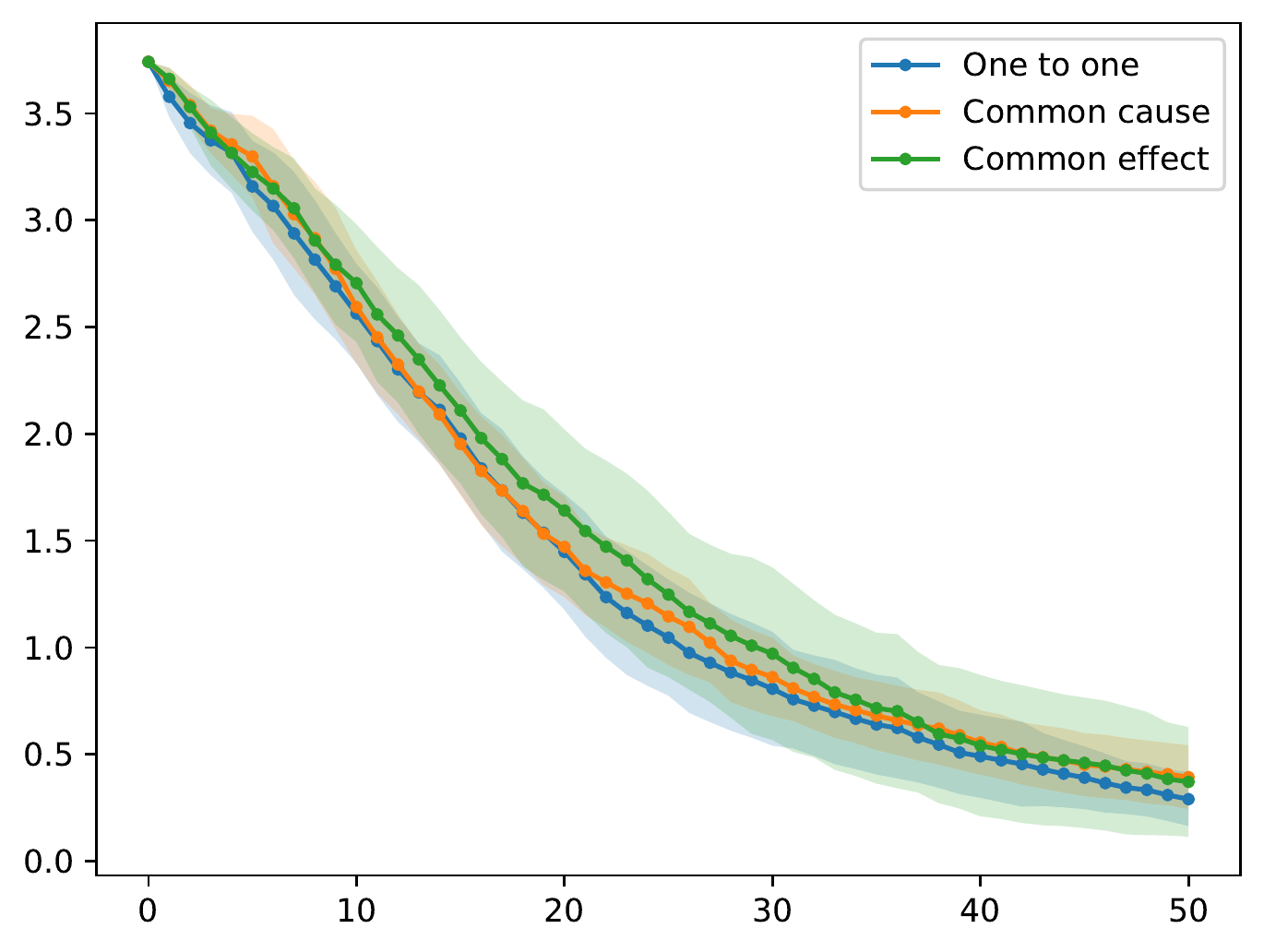}& \quad
\includegraphics[width=.2\linewidth]{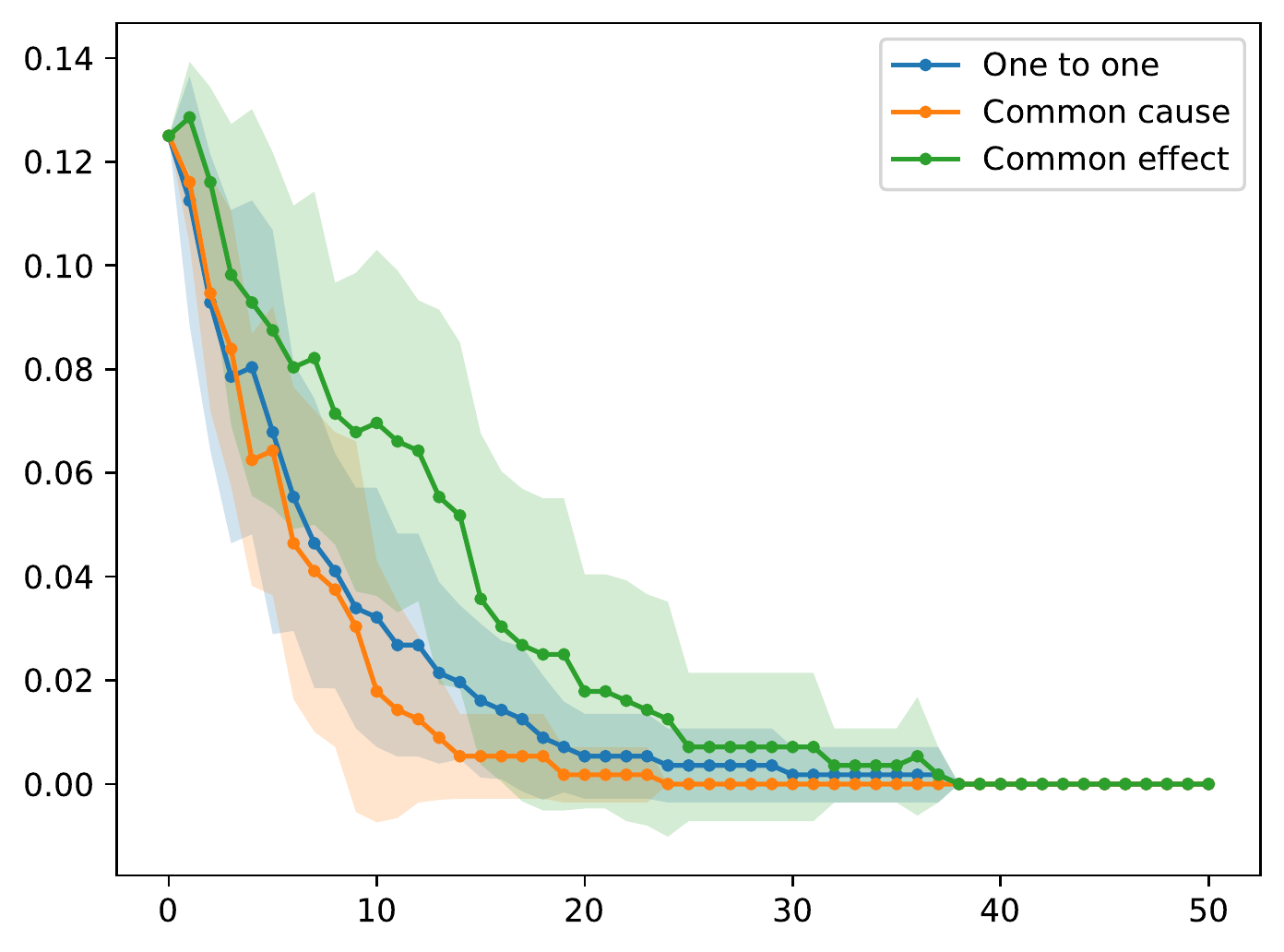}& \quad
\includegraphics[width=.2\linewidth]{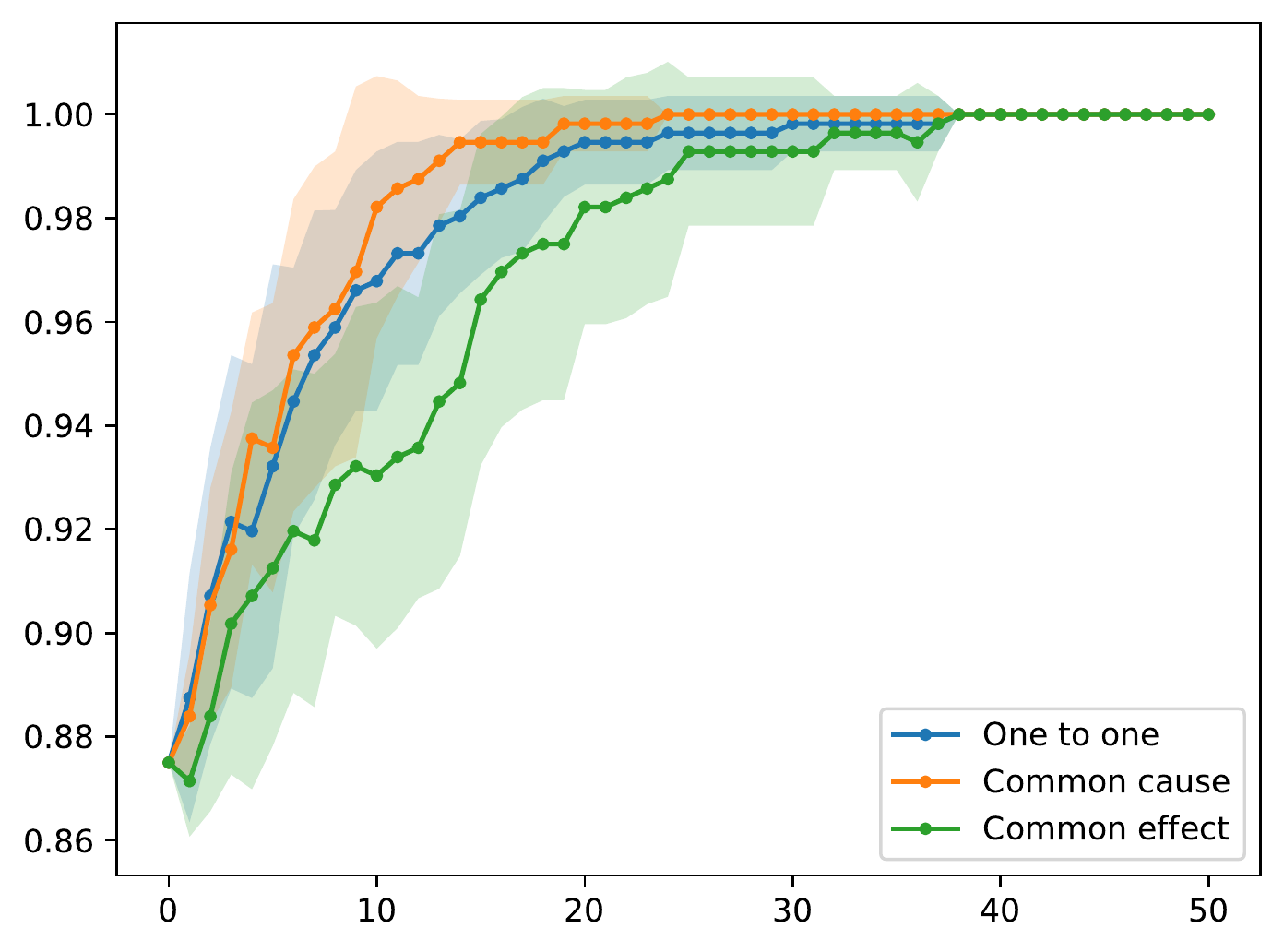}
\\
\rowname{\scriptsize{$N = 9$}}&
\includegraphics[width=.2\linewidth]{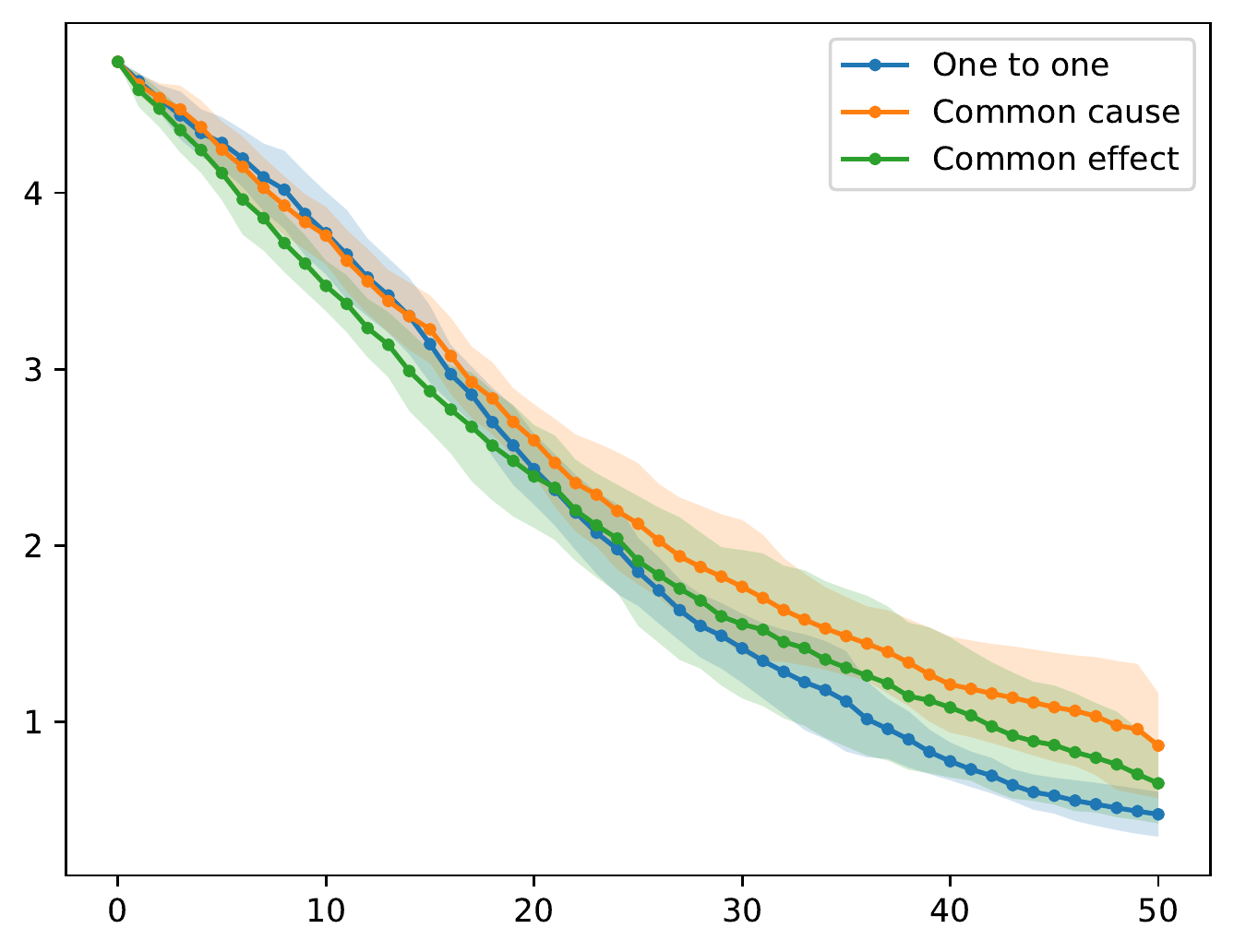}& \quad
\includegraphics[width=.2\linewidth]{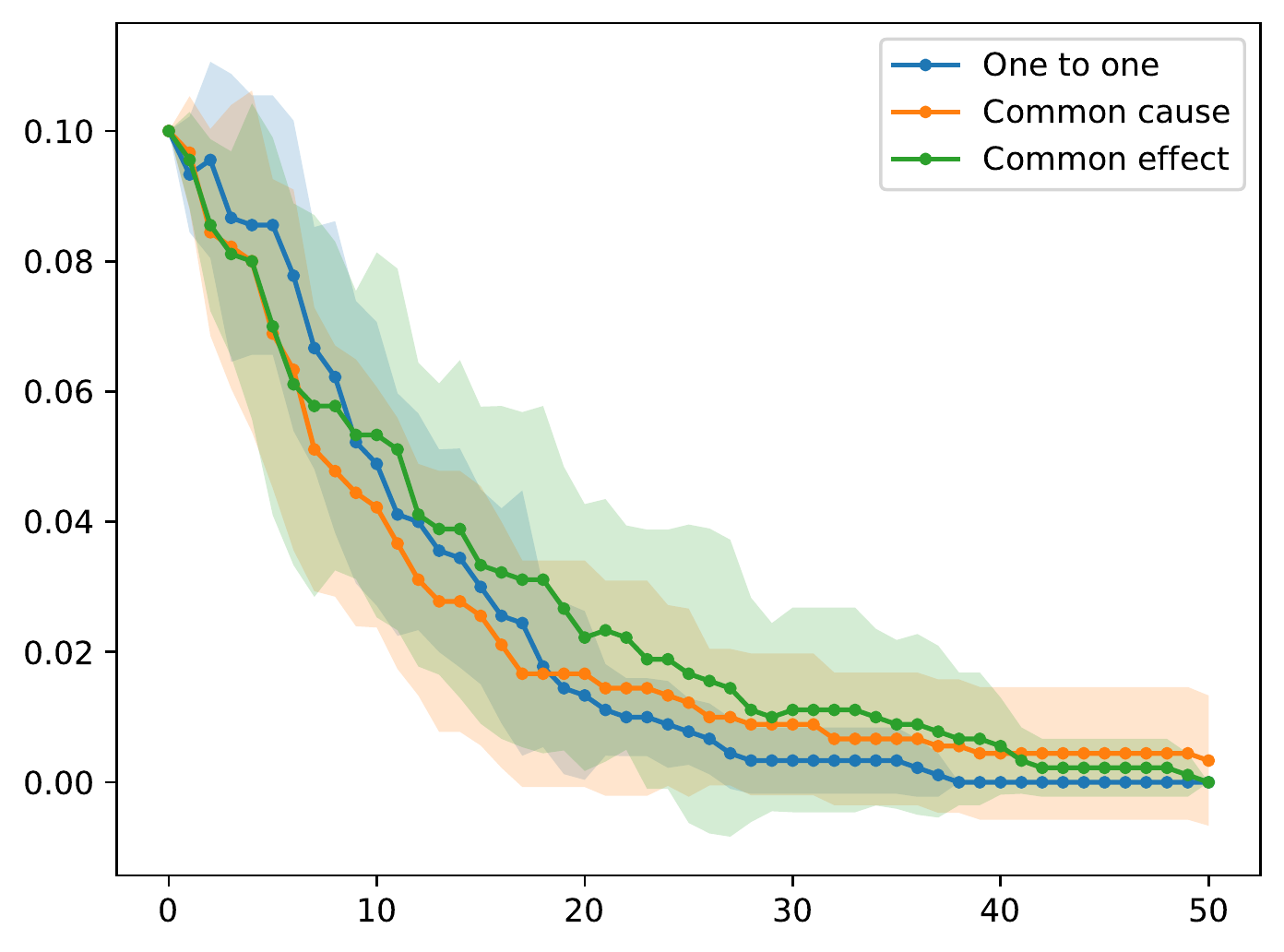}& \quad
\includegraphics[width=.2\linewidth]{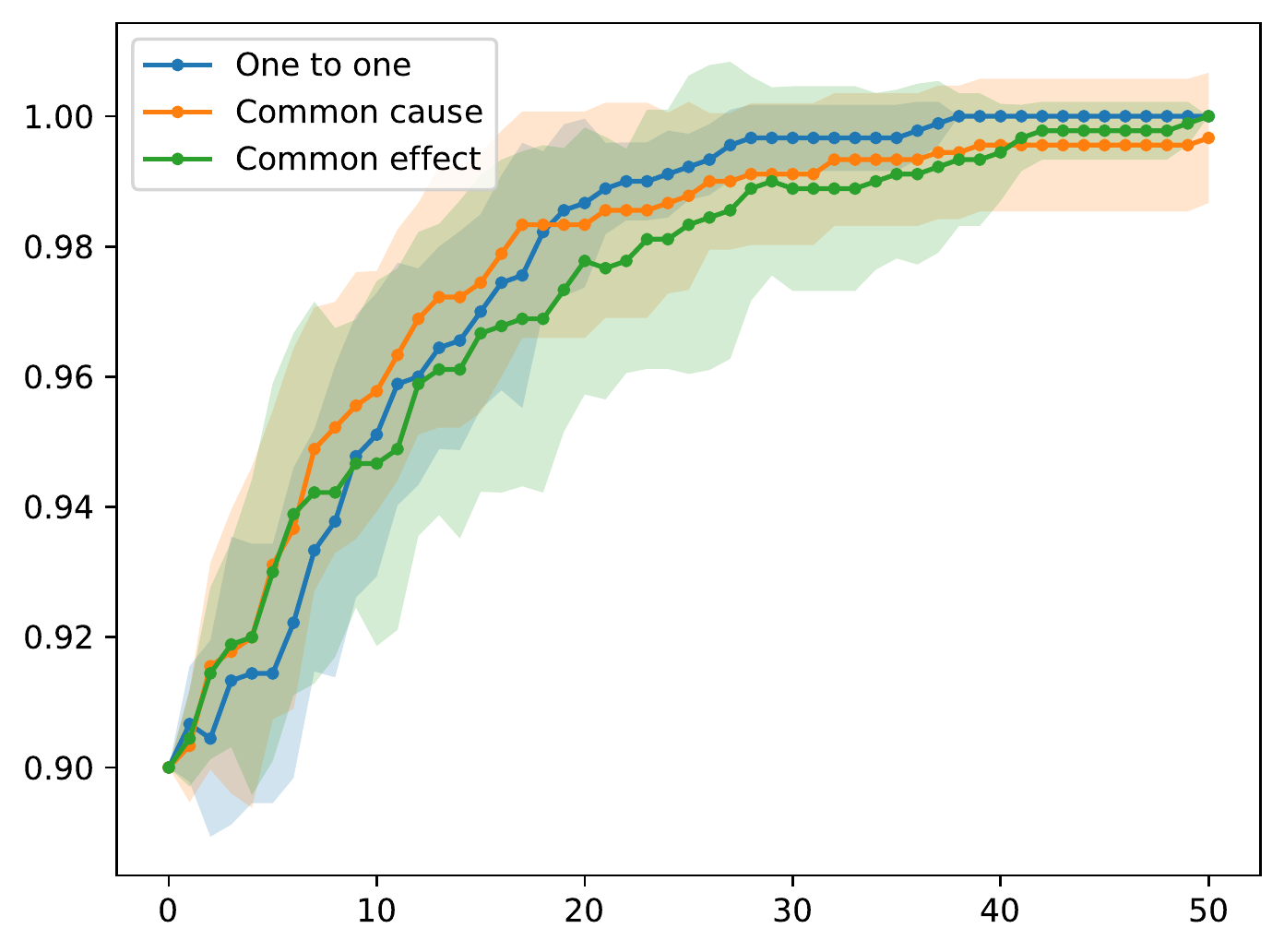}

\end{tabular}
\caption{Average value and standard deviation per round of each metric over $10$ runs.}
\label{fig:results-light-switches}
\end{figure}

Figure \ref{fig:results-light-switches} shows the evaluation metrics per $k$ rounds and the displayed value in the plots is the average over the $K$ different causal structures. Each plot corresponds to a number of lights $N$ and a different evaluation measure. We can see that the type of structures that represent a harder challenge are the \textit{Common effect} graphs. A possible explanation of this situation is that there is some ambiguity for the agent to know which switch causes a change in a light so there is a need to execute more rounds of interactions. In general, we can observe that the true causal seems to be learned for all the cases.

\begin{figure}[h]
\centering
        \begin{subfigure}[b]{0.3\textwidth}
                \centering
                \caption*{\scriptsize{One to one}}
                \vspace{-2mm}
            \includegraphics[width=0.89\linewidth]{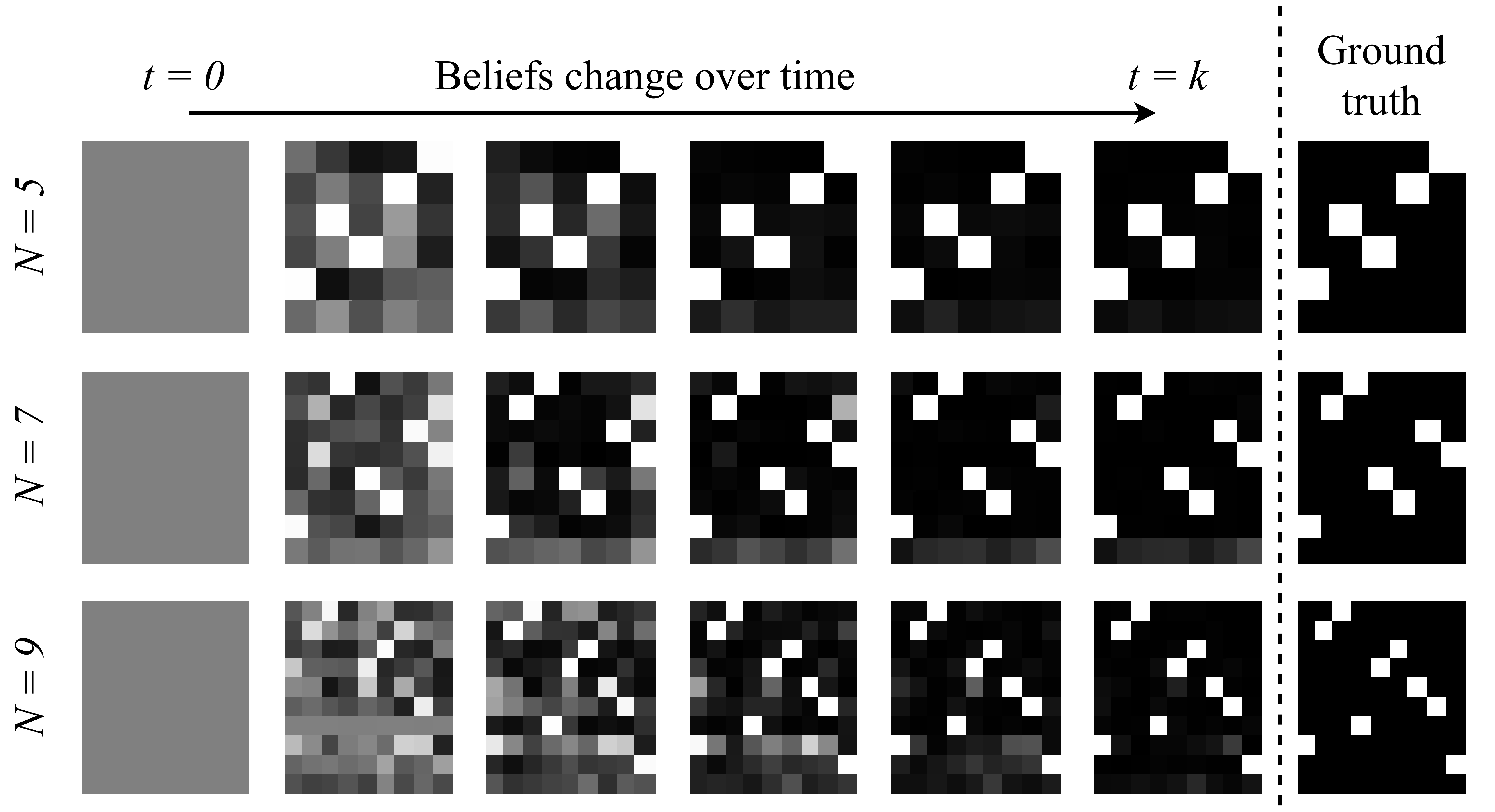}
        \end{subfigure}%
        \begin{subfigure}[b]{0.3\textwidth}
                \centering
                \caption*{\scriptsize{Common cause}}
                \vspace{-2mm}
               \includegraphics[width=.85\linewidth]{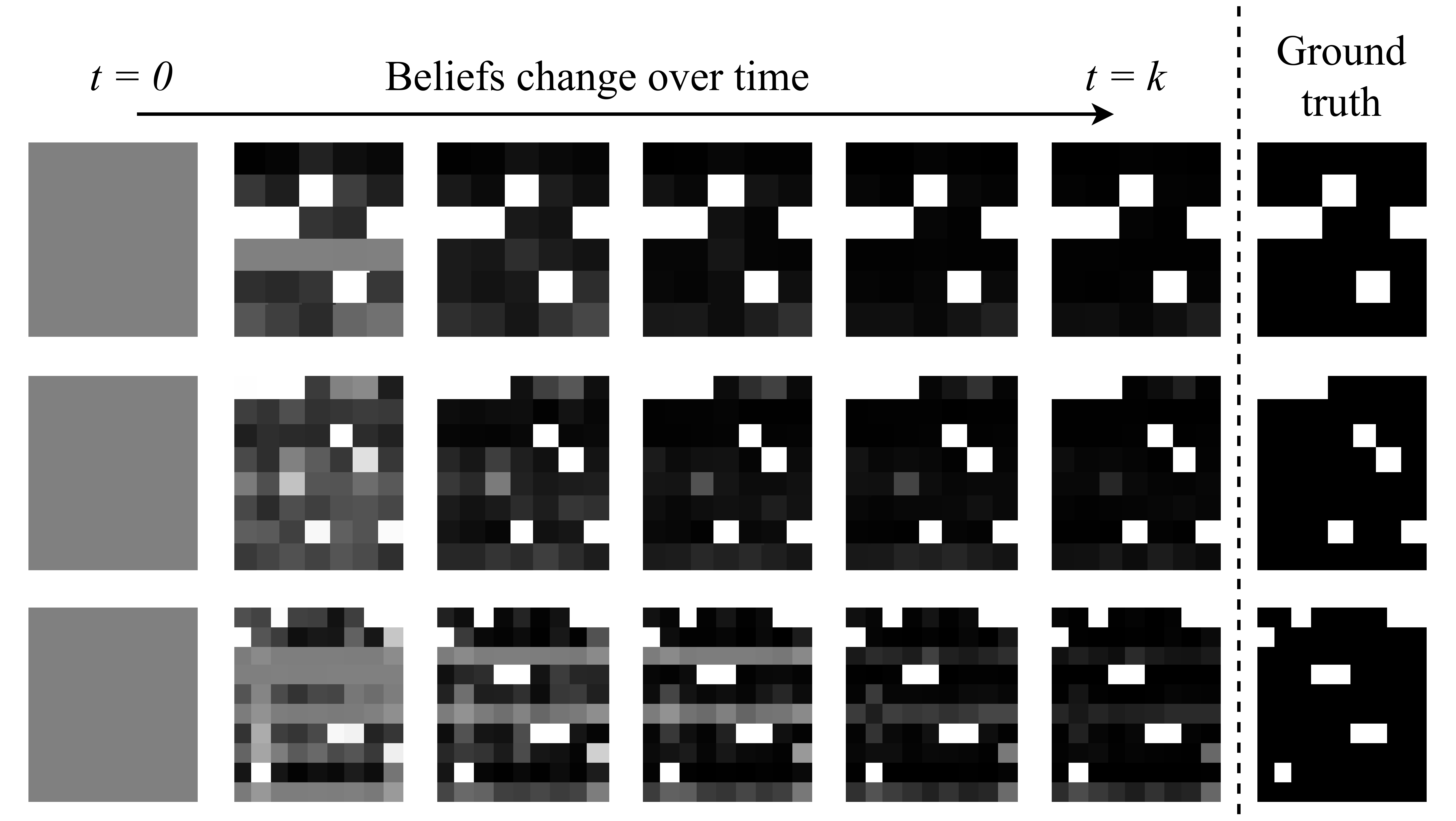}
        \end{subfigure}
        \begin{subfigure}[b]{0.3\textwidth}
                \centering
                \caption*{\scriptsize{Common effect}}
                \vspace{-2mm}
            \includegraphics[width=.85\linewidth]{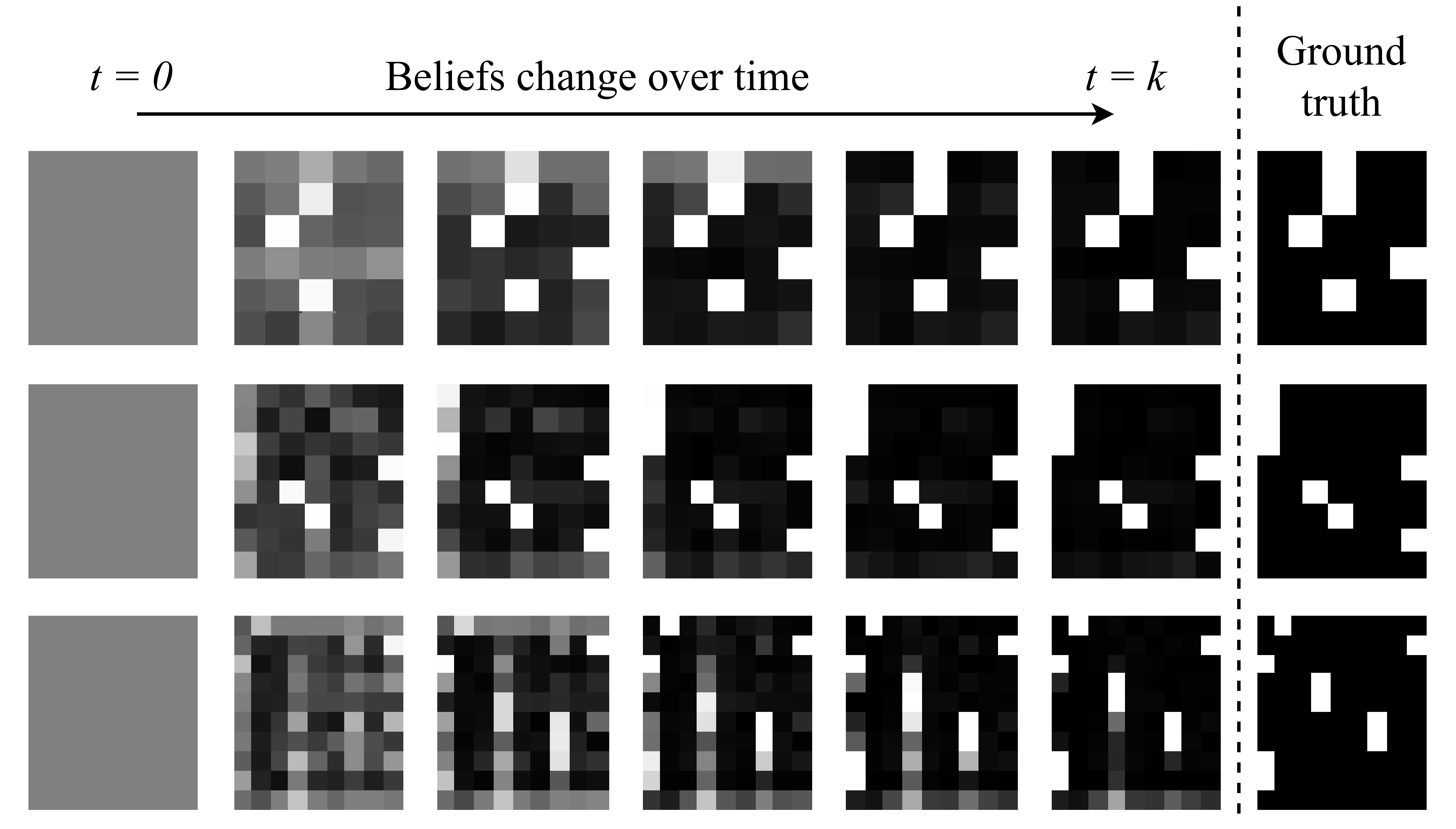}
        \end{subfigure}
        \caption{Example of heatmaps showing the changes over time of the beliefs against the ground truth, for the learning of one causal graph.}\label{fig:heatmap-light-switches}
\end{figure}

For better visualization of the change in beliefs. We randomly chose the updating process of one structure for every possible case. We collect the values of the beliefs at some learning time steps. Figure \ref{fig:heatmap-light-switches} shows the heatmaps describing the evolution of each belief. Analyzing the sequence of heatmaps, we can observe that the heatmaps obtained in the last beliefs updating are very similar to the ground truth in each case.

%\section{Future Work}
%In this work we have assumed the existence of a causal ordering over a known set of variables, which specifies which variables can not be a cause of other variables. In future work this assumption should be relaxed. We have considered an $\varepsilon$-greedy algorithm in order to balance exploration vs. exploitation in the disease-treatment scenario, we will later adopt such strategy in more complex scenarios. 
\section{Conclusions}
We have presented a Bayesian random graph-based methodology which is able to learn a causal structure from interventions and does not assume an initial DAG. Our approach is flexible and easily implementable as shown by the experiments, which also reflect a good performance over two quite different scenarios. In the first scenario we have also learned the optimal action while also learning a causal structure, which presented the challenge of balancing exploration and exploitation; we attacked this challenge using a $\varepsilon$-greedy algoritm with an exponential and linearly decaying $\varepsilon$.
\newpage
\bibliographystyle{apalike}
\bibliography{Arxiv_references.bib}
\end{document}